\documentclass[10pt,journal,compsoc]{IEEEtran}
\newif\ifpeerreview

\peerreviewfalse

\usepackage[nocompress]{cite}
\usepackage{url}
\usepackage{bbm}
\usepackage{amsmath,amssymb,graphicx, color}
\definecolor{cvprblue}{rgb}{0.21,0.49,0.74}
\usepackage[colorlinks=true, linkcolor=red, citecolor=cvprblue, urlcolor=cvprblue]{hyperref}

\usepackage{lipsum} %

\usepackage[switch]{lineno}
\usepackage{booktabs}
\usepackage{tikz}

\usepackage{graphicx}
\usetikzlibrary{positioning}

\newcommand{\paperID}{16}
\newcommand\C[1]\null
\title{Neural Ganglion Sensors:\\Learning Task-specific Event Cameras Inspired by the Neural Circuit of the Human Retina}

\author{Haley~M.~So
        and~Gordon~Wetzstein%
\IEEEcompsocitemizethanks{\IEEEcompsocthanksitem Haley M. So and Gordon Wetzstein are with the Department
of Electrical Engineering, Stanford University, Stanford,
CA, 94305, United States.\protect\\
\href{https://www.computationalimaging.org}{https://www.computationalimaging.org}}%
}

\begin{document}

\IEEEtitleabstractindextext{%
\begin{abstract}
Inspired by the data-efficient spiking mechanism of neurons in the human eye, event cameras were created to achieve high temporal resolution with minimal power and bandwidth requirements by emitting asynchronous, per-pixel intensity changes rather than conventional fixed-frame rate images.
Unlike retinal ganglion cells (RGCs) in the human eye, however, which integrate signals from multiple photoreceptors within a receptive field to extract spatio-temporal features, conventional event cameras do not leverage local spatial context when deciding which events to fire. Moreover, the eye contains around 20 different kinds of RGCs operating in parallel, each attuned to different features or conditions. Inspired by this biological design, we introduce Neural Ganglion Sensors, an extension of traditional event cameras that learns task-specific spatio-temporal retinal kernels (i.e., RGC ``events''). We evaluate our design on two challenging tasks: video interpolation and optical flow. Our results demonstrate that our biologically inspired sensing improves performance relative to conventional event cameras while reducing overall event bandwidth. These findings highlight the promise of RGC-inspired event sensors for edge devices and other low-power, real-time applications requiring efficient, high-resolution visual streams.
\end{abstract}

}

\maketitle

\IEEEraisesectionheading{
  \section{Introduction}\label{sec:introduction}
}
\IEEEPARstart{E}{vent} cameras can offer numerous advantages over frame-based image sensors that are crucial for the extreme constraints of emerging edge devices such as autonomous vehicles, robotics, and augmented/virtual reality. These include high temporal resolution, low latency, low power consumption, low bandwidth, and high dynamic range. Whereas traditional cameras output intensity frames at a fixed frame rate, a traditional event camera outputs asynchronous spikes that capture \textit{differences} in intensity. This design was initially inspired by the spiking nature of retinal ganglion cells (RGCs) which encodes the light hitting our retina into information-dense spikes and transmits the signals to our brain.  Event cameras have shown promise in many tasks including action recognition \cite{Plizzari_2022_CVPR, Gao2023, Ghosh2019SpatiotemporalFF}, optical flow \cite{Shiba23pami, Shiba22eccv, Tian_2022_BMVC}, depth or shape estimation \cite{Tulyakov_2019_ICCV, gu2022eventintensitystereo, Muglikar23CVPR}, and more \cite{eventreview}. However, event cameras are still limited compared to their biological analogues. While RGCs can aggregate information spatially across an area on the retina, event cameras only operate on a per-pixel basis: each pixel decides whether to send an event independent of neighboring pixels.

Most RGCs use local spatial information to decide whether or not to fire. There are roughly 20 kinds of RGCs in the human eye, and while the exact function of each varies, in general, these retinal ganglion cells receive information from not just a single photoreceptor, but rather a small group (as in midget cells) or from an even larger receptive field (as in parasol cells) to decide whether to send an action potential to the brain~\cite{wandell1995foundations}. One identified organization these RGCs operate with is a center-surround organization, in which the center receptive field is compared to the surrounding larger selected region to determine whether or not to fire a spike~\cite{kuffler1953discharge}. For example, CENTER ON cells look for where the center is on (light falls on the photoreceptor) and the surround is off, and CENTER OFF cells look for the inverse. In the human eye, this organization can allow for edge and contrast enhancement. In addition, there are direction-selective RGCs that are attuned to specific spatial frequencies, color-sensitive RGCs, and temporally attuned RGCs specialized for flicker or motion, to name a few. There are roughly 1.5 million RGCs in our eye that distill the information falling on our roughly 120 million photoreceptors, encoding the spatio-temporal information into sparse binary spikes. This raises our motivating question: How can we replicate the retina’s diverse retinal circuitry to achieve more efficient and versatile vision and perception?   

Towards this end, we revisit the retina’s bandwidth-efficient design principles and seek to learn the optimal set of functions for generating events, bridging the gap between conventional event cameras and their biological counterparts. Specifically, 
\begin{itemize}
    \item we introduce a framework for learning asynchronous, spatio-temporal events to optimize performance and bandwidth for perception tasks;
    \item we develop a differentiable event simulator;
    \item we show that integrating local spatial information can improve the performance on vision tasks over event cameras while exhibiting lower bandwidth;
    \item  we explore learning multiple complementary event ``channels,'' further enriching the captured information and boosting performance.
\end{itemize}

\section{Related Work}\label{sec:related}
\begin{figure*}[t]
    \centering
\includegraphics[trim={5mm 0mm 0cm 0mm}, clip, width=\linewidth]{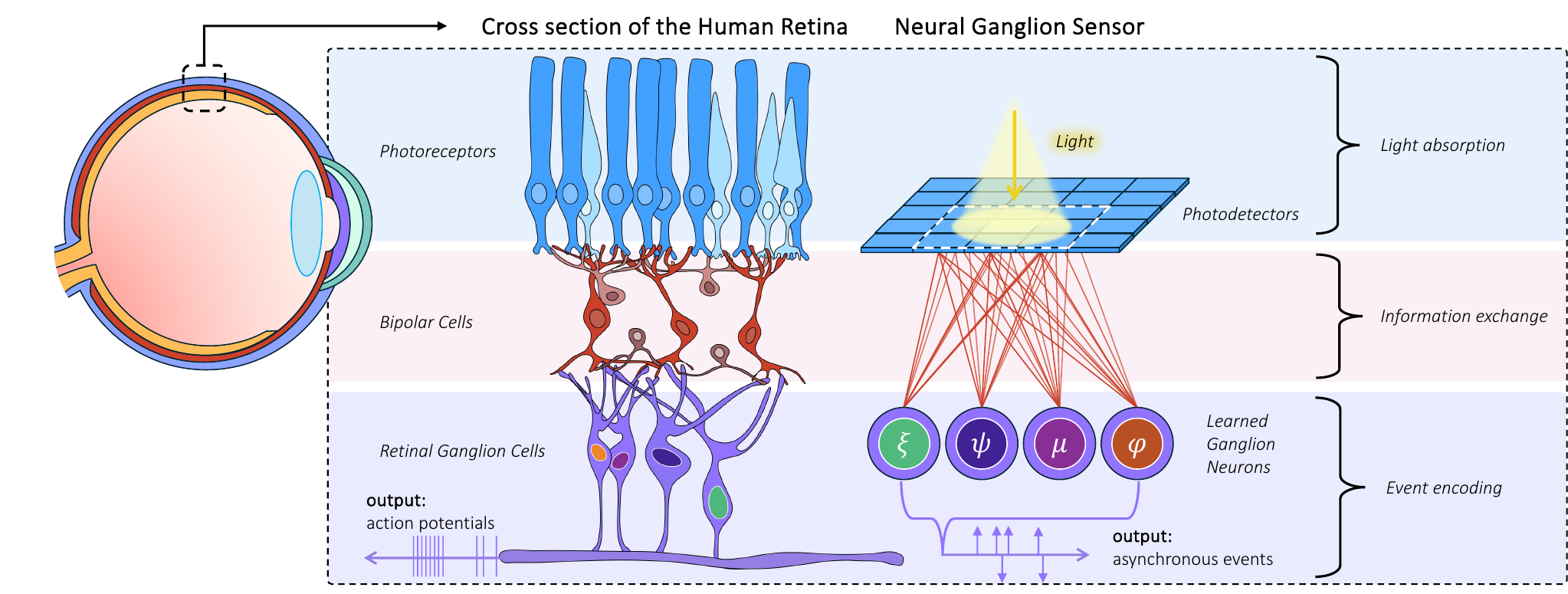}
    \vspace{-15pt}
    \caption{\textbf{Analogy between Neural Ganglion Sensors and the human retina}: On the left, we show a simplified diagram of different layers in the human retina. Light hits the photoreceptors (rods and cones), of which there are about 100 million per eye. The signals get transferred and modulated through Bipolar cells along with additional Horizontal and Amacrine cells. In the end, the roughly 1 million Retinal Ganglion Cells (RGCs), receive signals from a small \textit{area} on the retina, not just from a single photoreceptor. These RGCs look at the pattern of information to decide whether to send a spike signal to the brain. We see spatial and temporal pooling occurs in the first few layers of the retina to encode all the information into bandwidth efficient spiking potentials. On the right, we show our proposed Neural Ganglion Sensor, an event camera augmented to better match the human retina.}
    \label{fig:motivation}
    \vspace{-3mm}
\end{figure*}

\subsection{Retinal ganglion cells}
Recent literature has made significant progress in understanding how visual signals are processed even before they leave the human retina \cite{Chichilnisky2007, Guo2014}. Here, photoreceptors sense the incoming light, and the resulting signals are transmitted through and modulated by a diversity of horizontal, bipolar, and amacrine interneurons before being processed by roughly 20 different types of retinal ganglion cells (RGCs). The variation in interneuron pathways and RGC activation create multiple parallel retinal circuits or functions that extract a variety of complementary visual features at the retina \cite{Roska}, resulting in spikes sent to the brain. %
While the specific visual features identified by each RGC can vary, the output of RGCs can generally be described by similar structures. As a result, biologists often use simplified models such as the linear--nonlinear (LN) cascade model \cite{simplemodel, brackbill}, the Hodgkin–Huxley model \cite{HODGKIN199025}, and the various integrate-and-fire models \cite{Troyer1997, Jolivet, Pillow11003}. There are also recent works in using deep learning techniques to try to predict retinal responses to natural images \cite{deepretina, Botella-Soler, NIPS2017_b0169350}. From this literature, the key insight that we apply to our work is that ganglions receive signals not only across time but also across space from a receptive field of photoreceptors, not just a single one. In addition, our eyes also use multiple types of RGCs in parallel for efficient sensing.

\subsection{Event cameras}
Mahowald and Mead first introduced the silicon retina in the late 1980s \cite{mahowald_2008}, which quickly led to the development of the Dynamic Vision Sensor (DVS) or the event camera \cite{dvs-delbruck, PoschSpiking, Delbruck2016NeuromorophicVS}. In these systems, each pixel operates independently, always comparing the current intensity to a previously memorized intensity. If the intensity difference is larger than a set threshold value, an ``event'' is sent. The outputted information includes an x-y location, timestamp, and the polarity of the brightness change. The pixel then updates its memorized value to the intensity that triggered the output. Other variants of event cameras include the Asynchronous Time-based Image Sensor (ATIS) \cite{atis2011}, where an event trigger will also readout the intensity value at the given pixel or encode intensity through inter-spike time intervals \cite{boahen}, and the Dynamic and Active Pixel Vision Sensor (DAVIS) \cite{Berner2013A21, BrandliDAVIS}, which outputs full frame intensity values at a slow frame-rate along with the asynchronous events. While the events in these systems are inspired by the basic spiking nature in the retina, they don't utilize the spatial surroundings. Recently, Li and Delbr\"uck introduced the Center Surround Dynamic Vision Sensor (CSDVS) and illustrated the potential benefits of using a center-surround organization and suggested a future hardware implementation to achieve such a sensor with lateral polysilicon resistors and controllable transverse conductance \cite{Delbruck2022-csdvs, Li-thesis}. In 2023, \cite{svpevents} proposed using a pattern of different event thresholds across the sensor, analogous to spatially varying pixel exposures seen in computational imaging \cite{Nayar2000HighDR, neuralsensors, nguyen2022learning}. Most recently in 2024, \cite{Sundar2024} presented Generalized Event Cameras, which explored a few kinds of statistical ``differencing" methods, of which included the spatial dimension. They also introduced a nice breakdown of a general event camera as ``when to send" and ``what to send." While there are similar notes, the fundamental formulation as well as outputs are different: they output full intensity values while our approach remains truer to the human retina and the original event camera, sending just binary bits. Inspired by the human retina, we also uniquely explore learning multiple parallel task-specific RGCs and optimize specifically for bandwidth, not just performance.

\subsection{Event-based processing in vision applications}
There are a number of representations used to process asynchronous events in computer vision including event-by-event processing, events paired with intensity frames, and events processed in voxel grids.
Recent review papers \cite{eventreview, zheng2023deep} offer good insight into the advantages and disadvantages of each. Of these approaches, two of the most common ways to process events are either by utilizing spiking neural networks (SNNs) or by aggregating events into image-like frames and using conventional convolution neural networks (CNNs) \cite{Dalgaty}. 
While SNNs have been successful in a number of tasks including image classification \cite{zhou2023spikformer}, object detection \cite{su2023deep}, video reconstruction \cite{spikevideo} and more \cite{gu2020tactilesgnet, Amir, Botzheim, Zheng2023DeepLF}, they can be difficult to train as there is no traditional back propagation and they are relatively new, so are still catching up to the more mature field of CNN-based computer vision. As a result, most state-of-the-art networks \cite{Rebecq19pami, Maqueda2018EventBasedVM, Zhu-RSS-18, Deng2020, Zheng2023DeepLF} opt to aggregate events into voxel-like grids and use more traditional image-based computer vision techniques. As this is the most widely used method, we create a differentiable binning procedure to backpropagate through event voxel grids to be able to learn task-specific events.\\

In our work, we demonstrate that augmenting event cameras with additional biologically inspired spatial aggregation can improve performance for machine vision tasks. Furthermore, we explore learning multiple RGC channels and introduce a framework for optimizing these asynchronous events for bandwidth and performance.

\section{Proposed Method}\label{sec:methods}
\subsection{Model of RGC Events}
The linear-nonlinear cascade (LN) is a popular way to model the response of RGCs~\cite{simplemodel}. In these models, the Poisson spike rate of a neuron is determined by a linear spatio-temporal filter and a nonlinear activation. Specifically, the probability, $P$ that a neuron spikes can be described as a continuous 3D convolution of the intensity, $I$, over space $x,y$ and time $t$ followed by a non-linearity $f$:
\begin{align}
    P(x,y,t) &= f\biggl(\Bigl[W \ast I\Bigr]_{(x,y,t)}\biggr)
\end{align}
where $\ast$ denotes the convolution operation, and $W$ is a kernel with the weights of the spatiotemporal filter.

From the LN model, the event camera emerged with a few simplifications. 
Firstly, the probabilistic spiking is replaced with a deterministic activation if the output of $f$ exceeds a threshold $\delta$. Secondly, the continuous intensity $I$ is split temporally into the current intensity $I_{curr}$ and a memory intensity $I_{mem}$. Lastly, the $W$ filter is reduced to a simple temporal differencing. For Neural Ganglion Sensors, our RGC event formulation uses the first two simplifications of conventional event cameras, but reintroduces the spatial dimension. The resulting model for when a RGC event is triggered becomes:
\begin{align}
    P(x,y,t) &= \mathbbm{1} \biggl(f\Bigl(\Bigl[W \ast (I_{curr} - I_{mem})\Bigr]_{(x,y,t)}\Bigr)  > \delta \biggr)
    \label{eqn:rgc_trigger}
\end{align}
where $\mathbbm{1}$ is the indicator function.

Biological neurons are limited to outputting binary spikes, which is bandwidth efficient and fast to transmit. Similarly, when an event is triggered in event cameras, the output $O$ for each event triggered is binary and can be computed by the polarity of the difference:
\begin{align}
O(x,y,t) &= \begin{cases}
                \text{sign}(I_{curr}-I_{mem})_{(x,y,t)} &\text{if } P(x,y,t) = 1  \\
                \text{None} &\text{if } P(x,y,t) = 0 
            \end{cases}
            \label{eqn:output}
\end{align}
In addition, when the event is triggered at pixel $(x,y)$, $I_{mem}(x,y)$ is updated to $I_{curr}(x,y)$.
\begin{align}
    I_{mem}(x,y) &= \begin{cases}
                I_{curr}(x,y)   & \text{if } P(x,y,t) = 1 \\
                I_{mem}(x,y) & \text{if } P(x,y,t) = 0 
            \end{cases}
            \label{eqn:mem_update}
\end{align}
With these three equations defining the event trigger, the output, and the memory update, we have our full RGC event model. In fig.~\ref{fig:motivation}, we draw the parallel between the human retina and Neural Ganglion Sensors.

\subsubsection{Event Camera model}
Pixels in traditional event cameras operate independently of other pixels. 
Our RGC event formulation in equation~\ref{eqn:rgc_trigger} reduces to the traditional event camera when $W$ is simply the identity kernel, and $f$ is the absolute value function. 
\begin{align}
    P(x,y,t) &= \mathbbm{1} \left(\left| I_{curr} - I_{mem }\right|_{(x,y)} \geq \delta\right)
\end{align}
The equations for the output (eqn.~\ref{eqn:output}) and the memory update (eqn.~\ref{eqn:mem_update}) remain the same. 
 While this formulation enables event cameras to mirror the temporal aggregation of the retina, ganglion cells observe light from a receptive field, not just at a single point \cite{Roska}.

\subsubsection{Center--Surround Model}
One type of spatial aggregation for RGCs is the center--surround organization. At a high level, this allows for contrast or edge enhancements, spatial filtering, and more. In a given receptive field, the center pixels are compared to the surrounding pixels. Center ON cells look for when the center is excited and the surround is inhibited. Center OFF cells look for the inverse. Midget and parasol cells, two of the most ubiquitous cell types in the retina, each have Center ON and OFF configurations, though their spatial reaches differ, making them more attuned to different spatial frequencies. To achieve this spatial behavior with our RGC event formulation, the center pixel can be compared to the average value of the surround. $W$ would be set to a $k\times k$ kernel with $1$ in the center pixel and weights summing to $-1$ in the surrounding pixels to model a Center ON. Center OFF would be the same kernel, but negated.  Again, $f$ is the absolute value function. Similarly, another variation of the center--surround was suggested by \cite{Delbruck2022-csdvs}. In their case, the W kernel was set to: 
\begin{align}
   W= \begin{vmatrix}
    0 & 1 & 0 \\
    1 & -4 & 1\\
    0 & 1 & 0
    \end{vmatrix}
\end{align}

\subsubsection{Modeling other Human RGCs}
With our RGC model, we can model a number of other RGCs found in the human eye. Among vertebrates, orientation-selective neurons are attuned to edges in the cardinal directions (horizontal and vertical). In addition, some have selection for oblique orientations as well. To model these, we can use gradient-like filters or edge filters like the Sobel filter or even simpler binary filters. While the specific functions and the receptive fields of our 20 types of RGCs in our eyes are still being studied, the spatial dimension plays an integral role in event sparsification.

\subsection{Learning Task-Specific RGC events}
While the human retina inspires us, in the end, what visual features our eyes evolved to be attuned to may be very different from what machine vision would find important. As in deep learning where we moved from handcrafted features to learned features, here we follow a similar trajectory and learn what may be useful for perception tasks. We take our proposed model of a RGC event and learn the $W$ kernel and the threshold values to tailor our sensing for vision applications. In order to do so, we need to backpropagate through to the event generation, so we build a differentiable simulator which will be open-sourced.

\subsubsection{Simulating RGC events}
We construct our event generation based on ESIM and V2E~\cite{Hu2021v2e}, the two most widely used event simulators. However, we also support spatial kernels, allowing us to simulate diverse types of RGC kernels that goes beyond conventional events. We incorporate shot noise, non-uniform thresholds, refractory periods, separate and learnable positive and negative contrast thresholds, and the option to operate on log or linear intensity. Analogous to how human eyes have 20 kinds of RGCs, we also support learning multiple kinds of events, either through spatially-varying thresholds and kernels in a 2-by-2 bayer-like pattern or through multi-channel events. 

\subsubsection{Differentiable binning}
Current state-of-the-art event-based vision pipelines pre-process events into sparse frame-like images or sparse voxels. This allows researchers to build off of the plethora of works in frame-based computer vision. In this work, we seek to learn events for two different tasks, one that uses events and RGB images, as is common with the DAVIS sensor and other emerging industry sensors, and the other that uses just events. In both cases, the state-of-the-art works we build off of pre-process the events into a voxel-like data structure. To learn our RGC kernels, we must backpropagate through these voxel structures to the event generation, so we implement differentiable binning. Our binning is performed with a closed-form solution that can compute binned RGC events from high-speed video. This combines the functionalities of video-to-event simulators and conventional event pre-processing. Following recent approaches \cite{sun2023event, Wu_2024_ICRA}, events are weighted linearly into the two closest time bins. For each pair of frames at time $t_k$ and $t_{k+1}$ in the input high-speed video, the events generated between the frames are distributed into the two nearest neighboring bins at times $t_{\text{bin}}^-$ and $t_{\text{bin}}^+$, as described by the following equations. Here, $I_{RGC}$ is the output of the RGC kernels, $\alpha$ is the time spacing between events, $\beta$ is the time offset to $bin^-$, $pol$ is the polarity of the generated events, and $\mathcal{N}$ is the quantized number of generated events:
\begin{align}
    I_{RGC} &= W \ast (I_{curr} - I_{mem}) \nonumber\\
    pol &= \text{sign}(I_{RGC}) \nonumber\\
    \alpha &=  {\frac{t_{k+1}-t_k}{I_{RGC}/\delta}} \nonumber\\
    \beta &= (t_k-t_{\text{bin}}^-)  \\
    \mathcal{N} &= I_{RGC}//\delta \nonumber
\end{align}
where $\delta$ is either the positive or negative threshold, depending on the polarity of $I_{RGC}$.
Each event generated is weighted linearly into the two bins, depending on the distance, $w_i$, to each bin.  
\begin{align}
\begin{split}
    w_i &= \frac{\beta + {(i+1)}*\alpha}{t_{\text{bin}}^+ - t_{\text{bin}}^-} \nonumber\\
\end{split}\\
    \text{bin}^{-}_{i} & = \left( 1 - w_i \right) \cdot pol \\
    \text{bin}^{+}_{i} & = w_i \cdot pol \nonumber
\end{align}
Binning all the events generated between the pair of frames, we get the following:
\begin{align}
\text{bin}^{-}_{frame} &= \sum\limits_{i=0}^{\mathcal{N}-1} \text{bin}^{-}_{i} \nonumber\\
    \text{bin}^{+}_{frame} &= \sum\limits_{i=0}^{\mathcal{N}-1} \text{bin}^{+}_{i}
\end{align}
We can derive the closed-form solution for these summations by using the formula for the sum of an arithmetic sequence:

\begin{align}
\begin{split}
    \text{bin}^{-}_{frame} &= pol\cdot \left( 1 - \frac{\beta}{t_{\text{bin}}^+ - t_{\text{bin}}^-} \right) \cdot \mathcal{N}  \nonumber\\
    &- \left(\frac{\alpha}{t_{\text{bin}}^+ - t_{\text{bin}}^-} \right) \cdot pol\cdot \frac{(\mathcal{N}+1)(\mathcal{N})}{2}
\end{split}
\end{align}
\begin{align}
\begin{split}
    \text{bin}^{+}_{frame} &= pol\cdot \left( \frac{\beta}{t_{\text{bin}}^+ - t_{\text{bin}}^-} \right) \cdot \mathcal{N}  \\
    &+ \left(\frac{\alpha}{t_{\text{bin}}^+ - t_{\text{bin}}^-} \right) \cdot pol\cdot \frac{(\mathcal{N} +1)(\mathcal{N})}{2}
\end{split}
\end{align}

These equations are for a pair of frames and are applied for all the pairs in the video sequence to get the full voxel grid. With these closed-form equations and the straight-through-estimator~\cite{Bengio2013EstimatingOP} for quantized operations, we can now backpropagate from the binned events inputted to our vision models directly through to the RGC kernels, $W$, that we want to learn. We additionally extend our formulation to include non-zero refractory periods. See the supplement for these details.

\subsubsection{Optimizing Bandwidth and Performance }
In this work, we seek to learn the RGC kernels for two tasks: video interpolation and optical flow. We use task-specific loss functions, specifically Charbonnier pixel-wise loss and a masked L1 loss respectively. To optimize for \textit{sparsity} while fitting our learned RGC kernels, we add an additional weighted L1 loss on the number of events to push the model to learn RGC events that maximize performance while minimizing the number of total events.

\section{Experiments}\label{sec:experiments}
We experiment with two tasks to demonstrate the potential benefits of learning RGC events in both DAVIS (intensity + events) and DVS (events only) settings.
\begin{enumerate}
    \item Video interpolation is one of the most challenging tasks for event cameras and can act as an plug-and-play connection to perception tasks. The state-of-the-art uses output from a DAVIS camera.
    \item Optical flow is a particularly useful perception task. In this case, only events are used.
\end{enumerate}
For both tasks, we learn the RGC for improving performance and bandwidth. At the end of this section, we also delve into a number of questions the reader may be curious about, including if learning not one, but multiple kinds of RGC types, improves performance.

\subsection{Video Interpolation}
We train the state-of-the-art model \cite{sun2023event} by Sun et. al that performs video interpolation from events. In the original work, two intensity frames along with the events triggered in-between are fed into their model. From this, they reconstruct 7 frames in-between the two base frames. The original work used ESIM to simulate events from the high-speed GoPRO dataset ~\cite{Nah_2017_CVPR}. To learn our events, we replace ESIM with our differentiable simulator. We train with a variety of learned event settings and sparsity weightings to tune the overall average bandwidth. We train the interpolation model with a learning rate of $2\mathrm{e}{-4}$ and our RGC kernels with a learning rate of $5\mathrm{e}{-5}$ end-to-end for 200,000 iterations. We use the Charbonnier pixel-wise loss and our sparsity loss with varying weights to tune performance and bandwidth. Our simulator allows us to learn spatially-varying events too. In this setting, we learn a 2 by 2 bayer-like pattern of kernels and thresholds.

\begin{figure}
    \centering
        \includegraphics[trim={0cm 6cm 13cm 0cm}, clip, width=0.95\linewidth ]{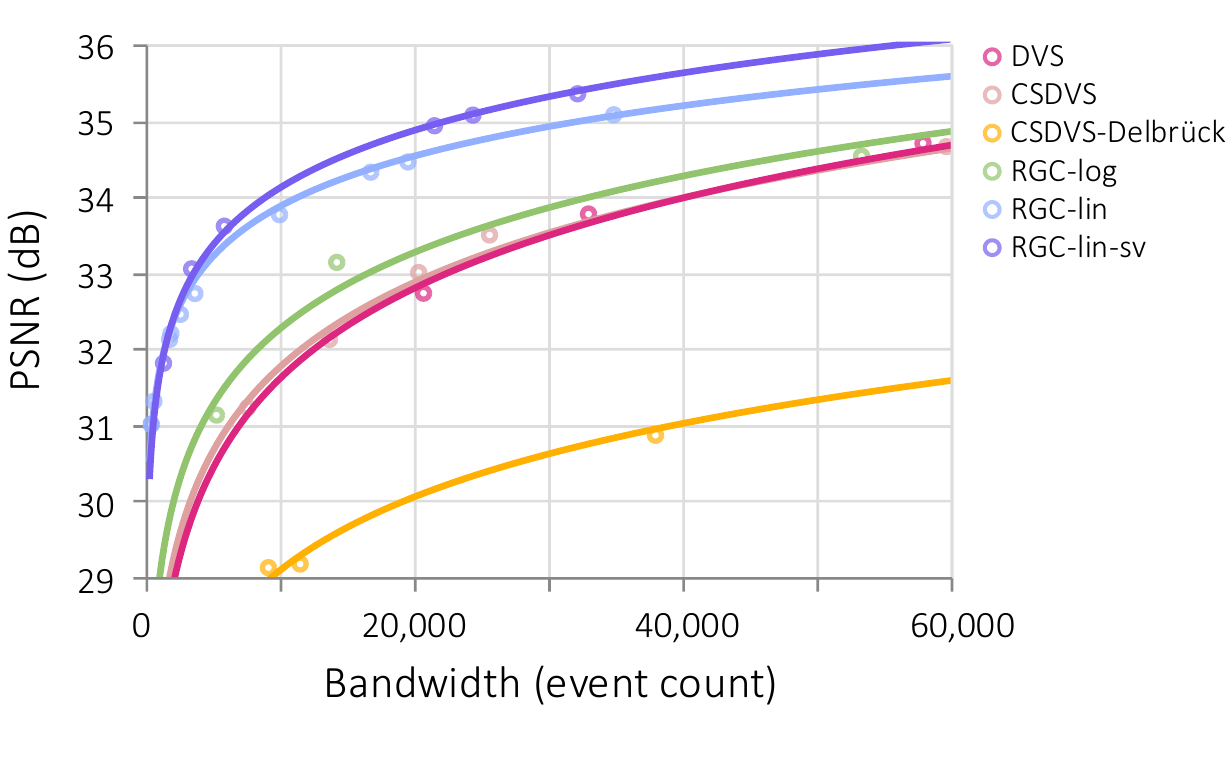}
    \vspace{-0.8cm}
    \caption{\textbf{Video Interpolation Performance vs Bandwidth Trade-off.} We perform video interpolation using DVS, CSDVS, CSDVS-Delbr\"{u}ck, RGC-log (learned, log regime), RGC-lin (learned, linear space), and RGC-lin-sv (learned, linear space, and spatially varying). For any given bandwidth, RGC-lin-sv provides the best performance. }
    \label{fig:7_interp}
    \vspace{-5mm}
\end{figure}

\subsection{Optical Flow}
For optical flow, we built off of the state-of-the-art optical flow from events model \cite{Wu_2024_ICRA} by Wu et. al. We utilize the TartanAir dataset \cite{tartanair2020iros} as it has ground truth optical flow. Similarly to \cite{Yang_2023}, we use EMA-VFI \cite{zhang2023extracting} to interpolate the RGB video before feeding the frames into our differentiable simulator, interpolating 15 frames between each pair. These intensity frames are used to simulate the RGC events but are not used to recover optical flow. In this task, only events are used to predict flow. We use the ``Hard'' subset of TartanAir. It provides 18 different scenes, with about 10 trajectories in each scene. Each trajectory has roughly 1,000 frames. To speed up training, we create train, validation, and test datasets from the ``Hard'' subset of the TartanAir Dataset, saving out random crops of the interpolated images and ground truth optical flow.  For maximal generalizability, we split train/val/test at the scene level. The generated dataset has 44,776 training sequences, 5,000 validation sequences, and 5,000 test sequences. Using this dataset and our differentiable event simulator, we learn the RGC kernels and optical flow model end-to-end.

\section{Results}\label{sec:results}
\subsection{Video Interpolation Results}
\begin{figure*}
    \centering

  \includegraphics[trim={0 5.5cm 0 0},clip,width=\linewidth]{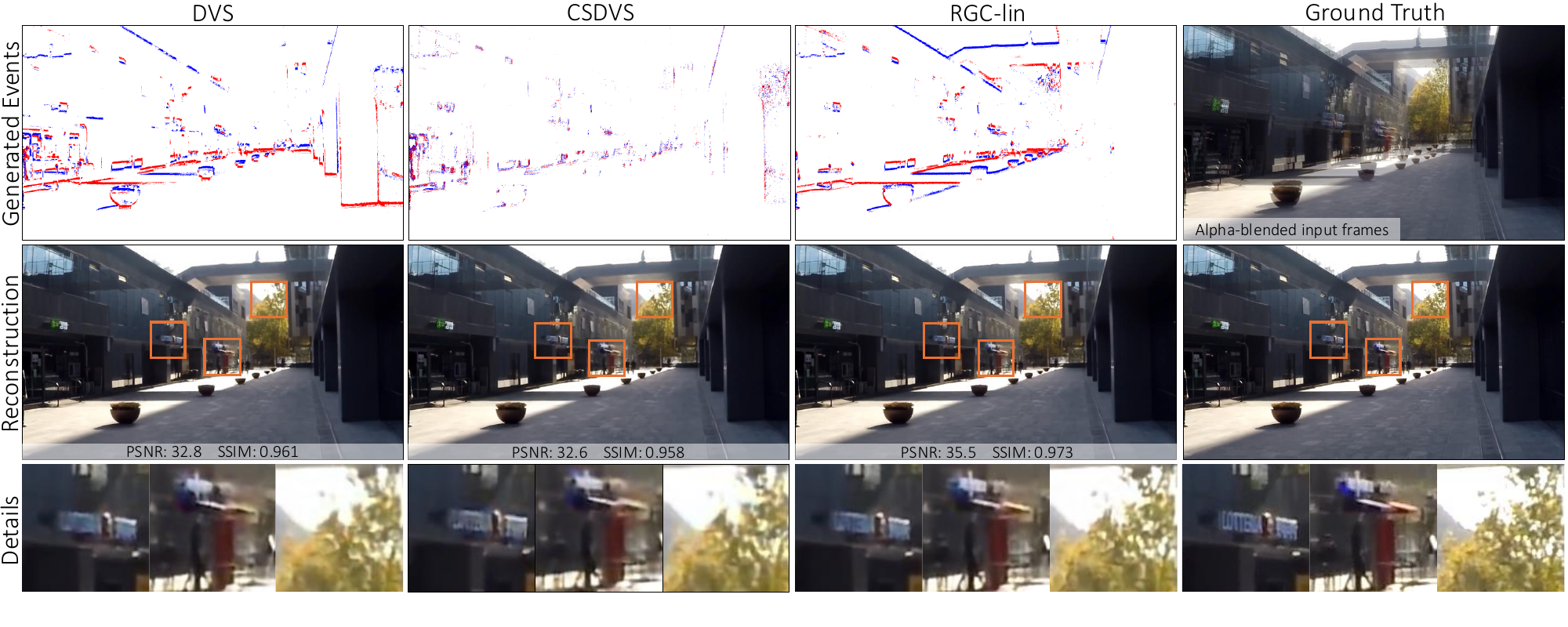}%
   \\
   \vspace{-3mm}
   \includegraphics[trim={0 5.5cm 0 0},clip,width=\linewidth]{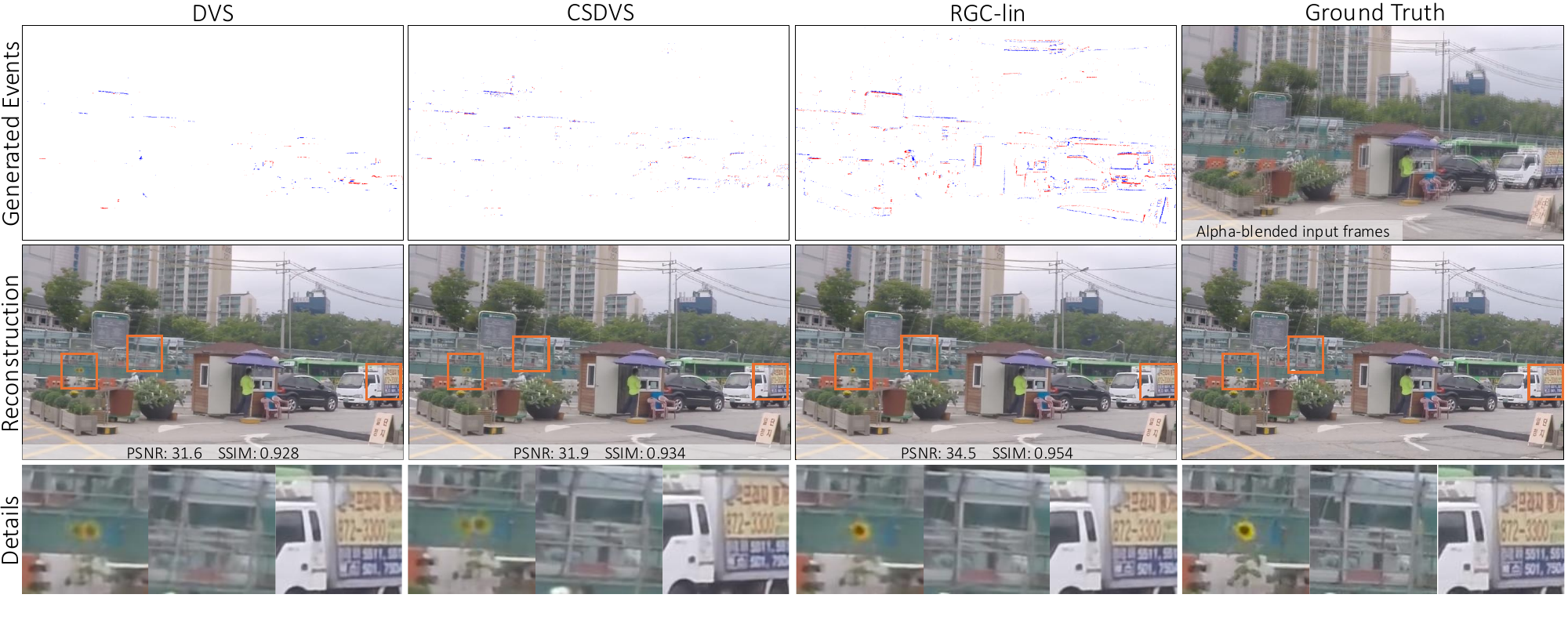}%
\\
\vspace{-3mm}
\includegraphics[trim={0 5.5cm 0 0},clip,width=\linewidth]{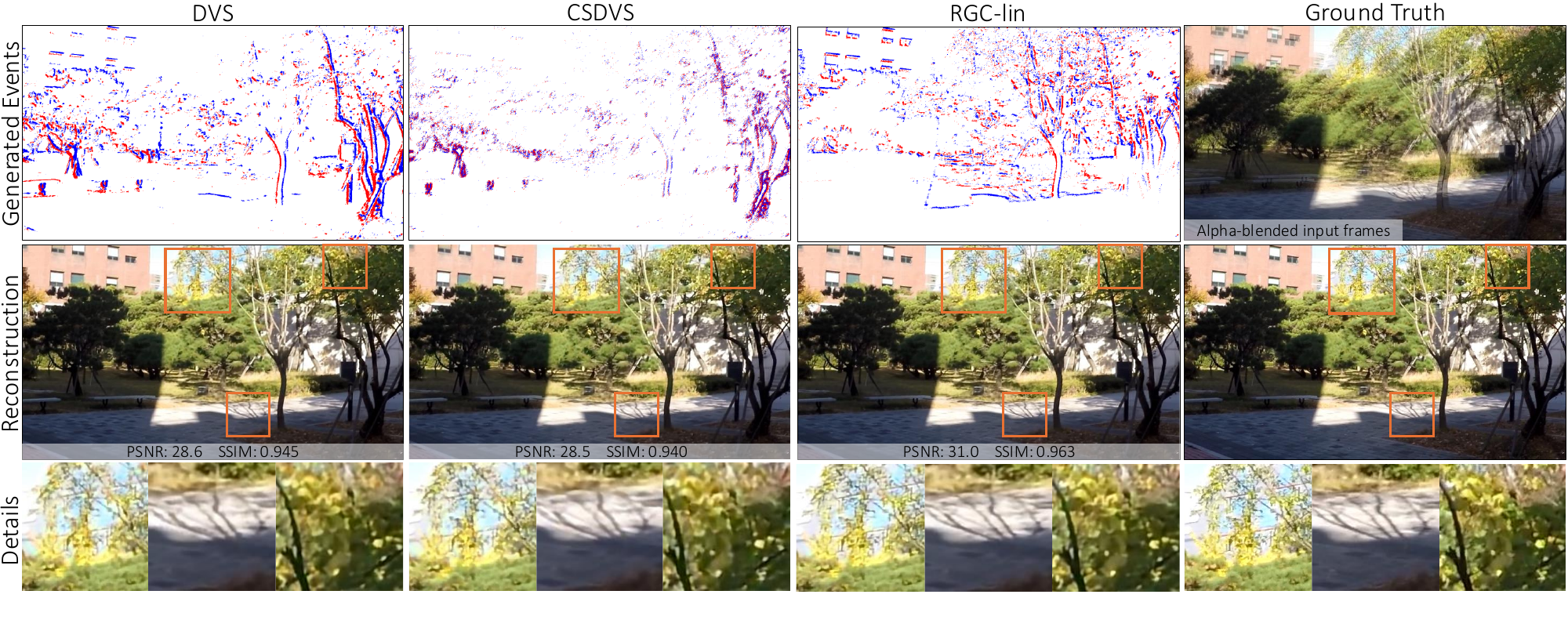}%
\vspace{-5mm}
\caption{\textbf{Video Interpolation Qualitative Results.} For each scene, we compare the reconstructions of the middle frame in the sequence for DVS, CSDVS, and RGC-lin. The top row shows the generated events, binned into the corresponding middle time bin, the second is the predicted image and the bottom row shows zoom-ins. The right-most column shows the start and end frames, alpha-blended, ground truth frame, and zoom-ins. PSNR($\uparrow$) and SSIM($\uparrow$) metrics are shown for each reconstruction.}
\label{fig:interp_grid}
\vspace{-3mm}
\end{figure*}

We present the bandwidth vs. performance tradeoff for interpolating 7 frames between pairs of frames, recovering 240fps from 30fps. Fig. \ref{fig:7_interp} shows the trade-off between bandwidth and performance of different types of events including the traditional DVS, the two variants of the center-surround (CSDVS and the hand-crafted $\text{CSDVS}_{\text{Delbr\"{u}ck}}$), our RGC-log (learned RGC in the log intensity domain) and RGC-lin (learned RGC in linear domain). We also show spatially-varying events in the same plot as RGC-lin-sv. 

As shown, RGC-lin achieves better performance at any given bandwidth compared to the traditional DVS. Furthermore, adding spatially varying events and thresholds pushes the performance vs bandwidth pareto front up to the left even more. For example, DVS achieves 33.8dB at 33.0k average events per bin while RGC-lin-sv achieves 35.4dB with 32.2k events, a 1.6dB increase using the same number of events. Similarly, if we look at a given performance, such as the 33.8dB that DVS achieves in 33.0k events, RGC-lin achieves 33.8dB with 9.9k events or over $3.3\times$ fewer events, and RGC-lin-sv achieves 33.6dB with 5.9k events or over $5.7\times$ fewer events.  These clear benefits highlight the promise of our learned RGC events.

Fig. \ref{fig:interp_grid} shows a few examples of the reconstructed videos. The comparisons are for a given average bandwidth of about 20,000 events per bin, specifically $20,716$ events for DVS, $20,357$ for CSDVS, and $19,557$ for RGC-lin. Averaging over the full GoPRO test set sequences, DVS achieves 32.7dB in PSNR, CSDVS achieves 33.0dB, while RGC-lin reaches 34.4dB, 1.67dB higher than DVS. As the middle frame in the sequence is the most challenging to predict, images and metrics (PSNR and SSIM) shown are for the middle frame in each scene. We show the events generated and zoom-ins to details in the reconstructed images. In the ground truth column, we also show the alpha-blended start and end frames of the sequence to illustrate the amount of motion being interpolated. Quantitatively, our learned RGC-lin achieves higher PSNR and SSIM given the same bandwidth as DVS or CSDVS. Qualitatively, the reconstructed structures are sharper and truer to the ground truth images.

\subsection{Optical Flow Results}
\begin{table}[]
    \scriptsize
    \caption{\textbf{Optical Flow Quantitative Results. } We compare models trained on DVS with different contrast threshold magnitudes, (which effectively changes the bandwidth), against two of our learned RGC-lin models at different bandwidths. $\text{RGC-lin}_{\text{lite}}$ and RGC-lin are the same base model, just trained with different sparsity loss weightings. End-point-error (EPE) is the L2 norm between the predicted and ground truth flow. 1PE is the percentage of pixels that have a predicted flow that is off by more than 1 pixel. 3PE is the percentage of pixels off by more than 3 pixels.}
    \centering
    \begin{tabular}{lccccc}
    \toprule
         & DVS $0.1$T & DVS $0.3$T & DVS $0.5$T & $\text{RGC-lin}_{\text{lite}}$ & RGC-lin  \\
         \midrule
         Bandwidth & $7.30$M & $4.11$M & $2.77$M & $2.10$M & $3.80$M\\
         $\downarrow$ EPE & $2.80$ & $3.02$ & $3.33$ & $2.75$ & $\mathbf{2.42}$\\
         $\downarrow$ 1PE & $55.1$ & $60.8$ & $66.3$ & $52.4$ & $\mathbf{47.1}$ \\
         $\downarrow$ 3PE & $20.1$ & $23.0$ & $26.1$ & $20.7$ & $\mathbf{17.4}$\\
     \bottomrule
    \end{tabular}
    \label{tab:flow_metrics}
    \vspace{-3mm}
\end{table}

In this task, solely events are used to reconstruct the optical flow. Similarly to interpolation, learning the kernels provides improved performance and lower bandwidth. In Table~\ref{tab:flow_metrics}, we compare the models trained end-to-end with our learned RGC-lin to multiple models trained on DVS outputs of different contrast thresholds. The higher the contrast threshold, the fewer the events. RGC-lin and $\text{RGC-lin}_{\text{lite}}$ are the same model, just trained with different sparsity weightings resulting in different average bandwidths.  Over the test set, our learned kernel for optical flow provides the best performance over all metrics. We use the standard metrics: End-point-error (EPE) is the L2-norm between predicted and ground truth flows, 1PE is the percentage of pixels whose flow is off by more than 1 pixel, and 3PE is the percentage of pixels whose flow is off by more than 3 pixels. For DVS, the performance increases with the number of events. However, the best performance comes from the learned RGC-lin kernel, which achieves an EPE of 2.42, better than even the DVS model with nearly twice the number of events. In fact, at the same performance as the DVS 0.1T model, the learned approach only needs 2.10M events, which is 3.5 times fewer events.

\begin{figure}[t]
    \centering
    \includegraphics[width=\linewidth]{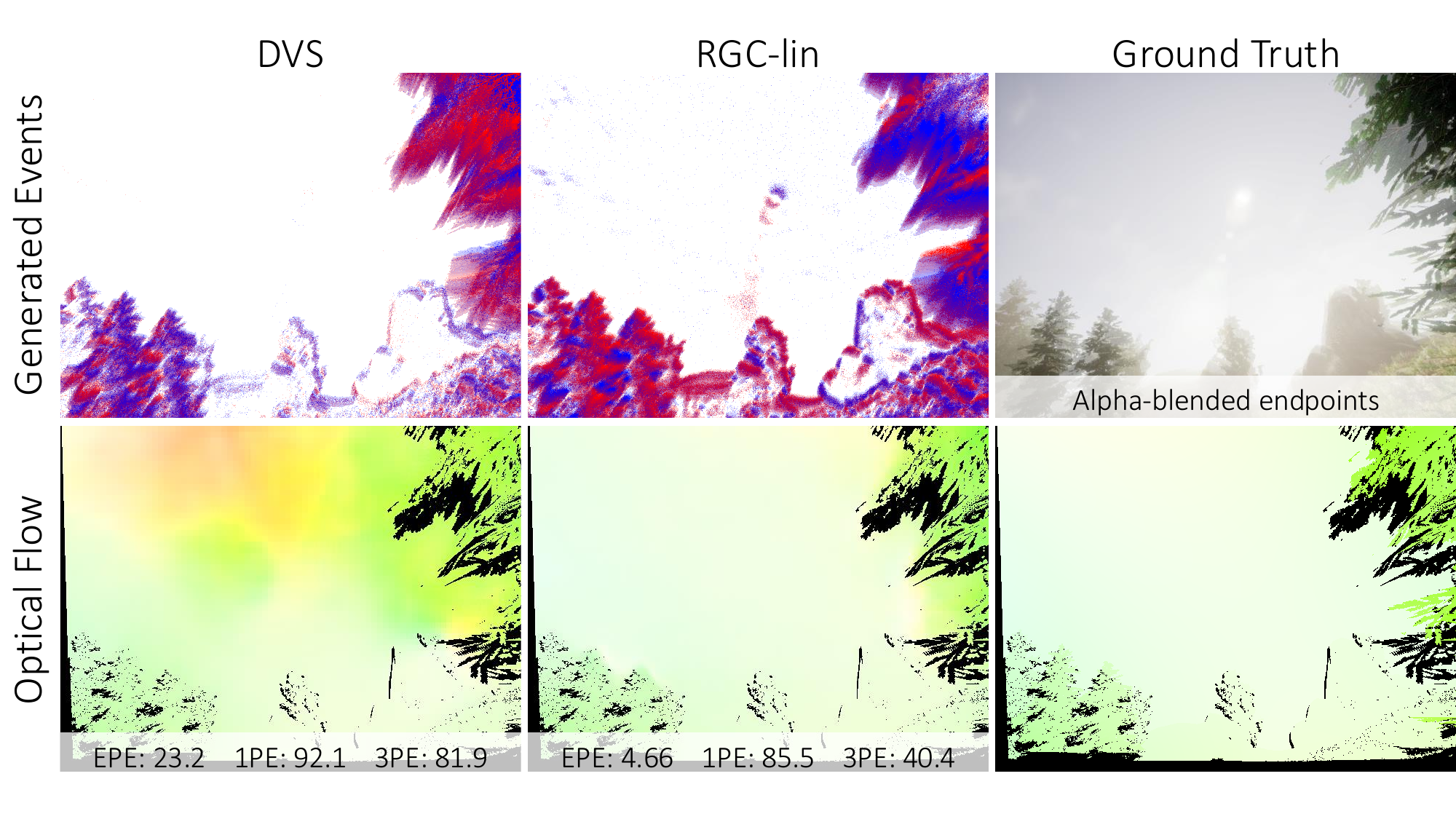}
    \\
    \vspace{-2mm}
    \includegraphics[width=\linewidth]{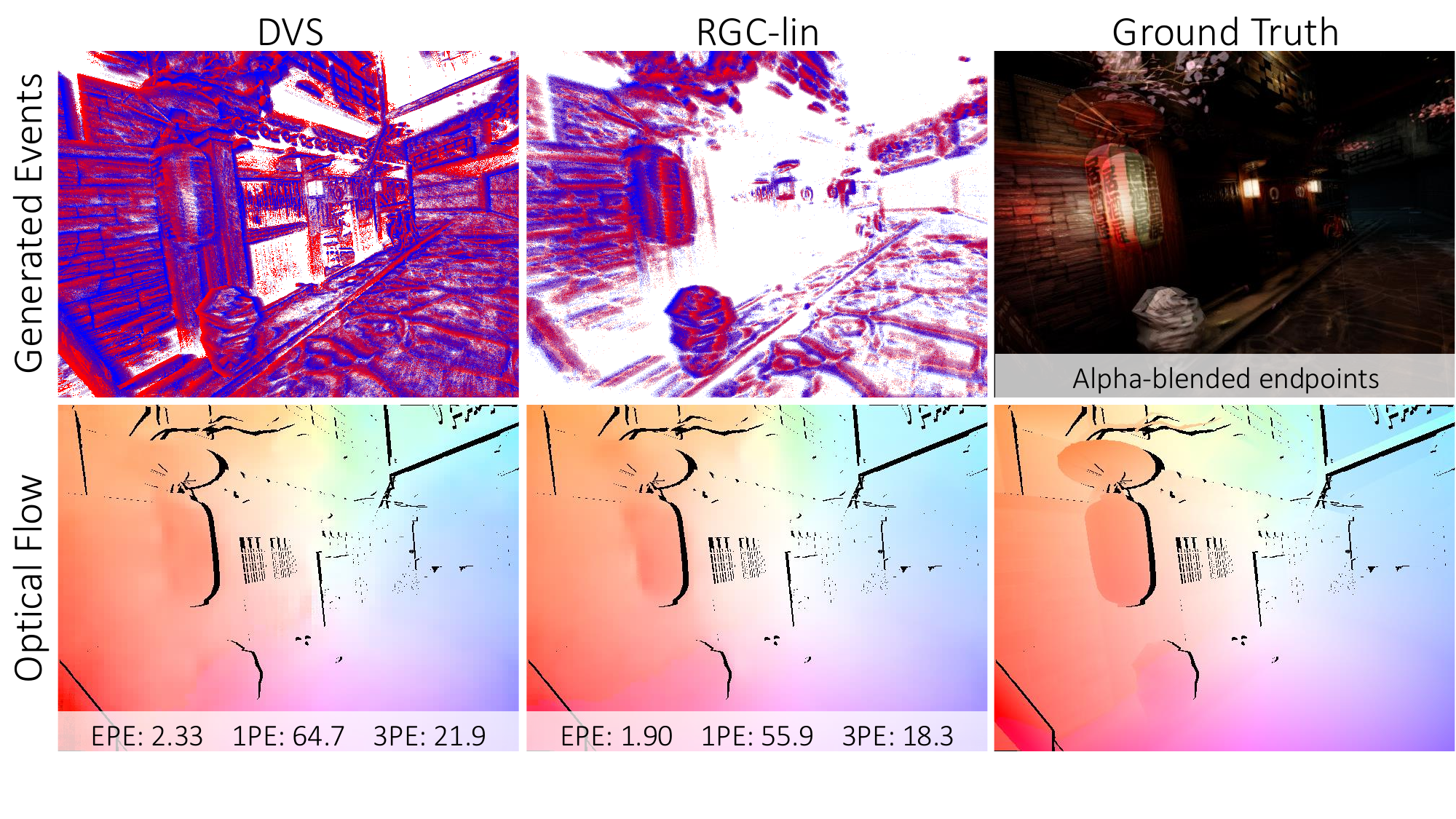}
    \\
    \vspace{-2mm}
    \includegraphics[width=\linewidth]{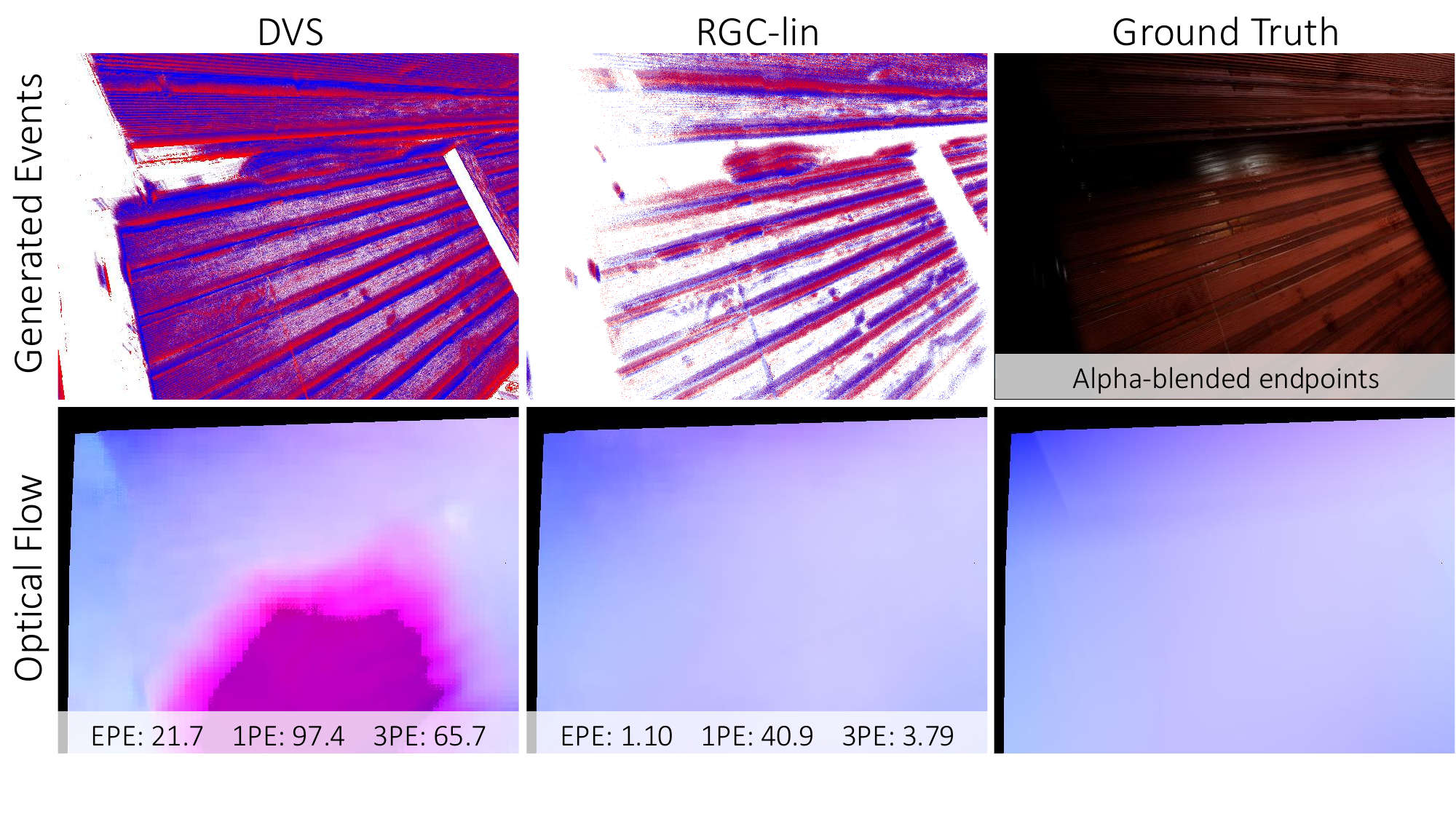}
    \vspace{-9mm}
    \caption{\textbf{Optical Flow Qualitative Results.} For each sample, the top row shows the events generated by the DVS kernel and our learned RGC-lin kernel as well as the alpha-blended camera frames \textit{just for reference}. In this task, solely events are used to reconstruct the flow. The bottom row shows the reconstructed flows and the ground truth flow.  We show EPE$\downarrow$, 1PE$\downarrow$, and 3PE$\downarrow$ metrics for the reconstructions. }
    \label{fig:flow_results}
\end{figure}

Fig. \ref{fig:flow_results} shows three samples of the optical flow reconstruction from DVS and from our Learned RGC-lin events. The DVS model corresponds to DVS 0.1T from tab.~\ref{tab:flow_metrics}, as it had the best performance among the DVS models. Learning the kernel end-to-end allows RGC-lin to better reconstruct the flow of the whole frame than DVS while lowering overall bandwidth.

\begin{figure}[t]
    \centering
    \includegraphics[trim={0 10.5cm 1.5cm 0},clip,width=\linewidth]{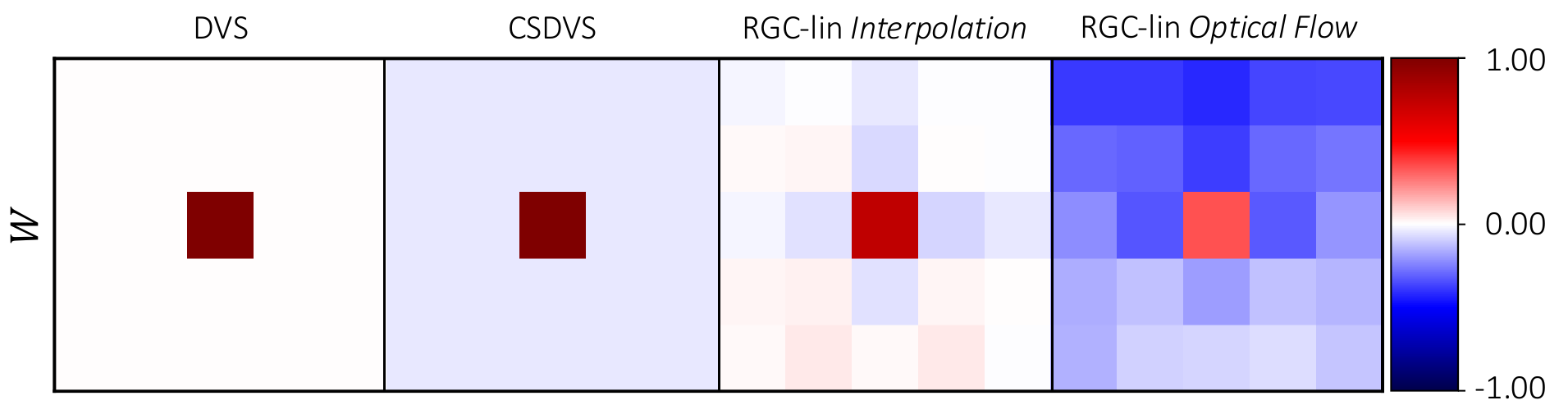}
    \vspace{-7mm}    \caption{\textbf{Comparison of Kernels.} We show the $5\times5$ kernels for DVS, CSDVS, the learned RGC-lin kernels for video interpolation, and the learned RGC-lin kernels for optical flow.}
    \label{fig:kernel_comparison}
\end{figure}

\subsection{Insights from the Learned Kernels}
In fig.~\ref{fig:kernel_comparison}, we show a comparison of the DVS, CSDVS, and the learned kernels. As we can see, the best kernels for interpolation and optical flow differ greatly, which is not surprising as the tasks are quite different. However, this highlights how choosing a single sensing kernel like the DVS for every task can limit the potential performance. 

For interpolation, the learned kernel reveals that the model places importance on local contrast as its ring structure emphasizes sharp changes in intensity. It effectively captures something akin to an edge detection or bandpass filter, which can be a powerful cue for figuring out how intermediate frames should look. 

For optical flow, the kernel has even greater contrast. While the kernel for interpolation was nearly symmetric, the kernel for optical flow is strongly asymmetric, exhibiting some direction-selectivity or the importance of directional gradients. In our eyes, we also have direction-selective RGCs that are more attune to detecting motions, and they are also asymmetric or have elongated receptive fields.

Additionally, in our supplement, we study how the learned kernels change as we increase the receptive field and as we increase bandwidth for video interpolation. In the first study, we sweep kernel sizes $k=3,5,7,9,11$ and find that the performance stays roughly the same at a given bandwidth. In the second study, we increase the weighting on the sparsity loss to tune the bandwidth. We include visualizations of the corresponding kernels in the supplement.

\subsection{Multi-Event Exploration}
\begin{figure}[t]
    \centering
    \includegraphics[trim={0, 0cm, 17.5cm, 0}, clip, width=\linewidth]{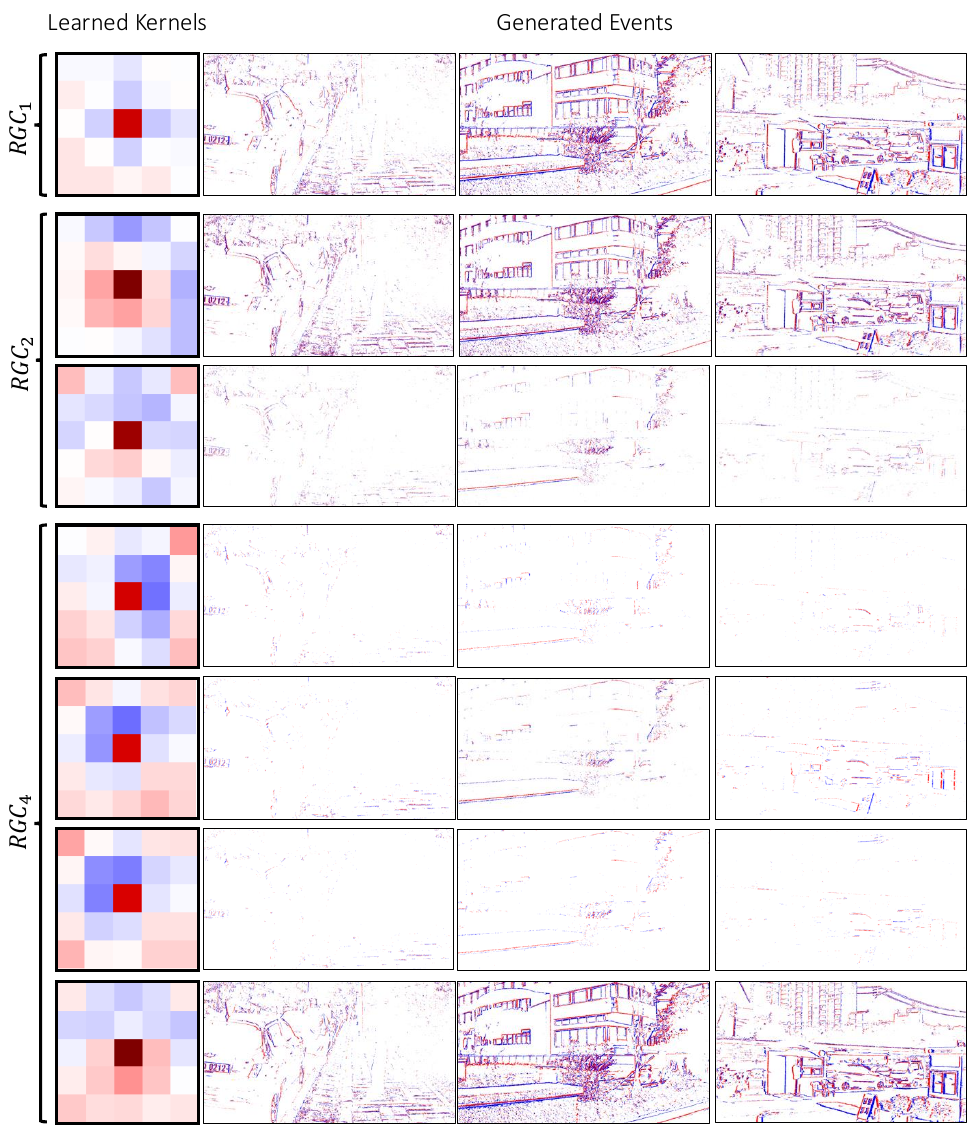}
    \caption{\textbf{Comparison between learned single and multi-event.} We compare the learned kernels of a single RGC model, two RGC, and four RGC for interpolation and show the generated events for each. Each column is a different scene At the same bandwidth, the 2-kernel model had a $0.40$dB improvement over the single kernel and the 4-kernel model had an additional $0.35$dB over the 2-kernel model.  }
    \label{fig:multiE}
    \vspace{-3mm}
\end{figure}

Inspired by the diversity of RGCs in the human eye,
we explore models with increasing number of learned RGC types per pixel location for the task of interpolation. With one RGC type, the kernel is roughly symmetric, but as the number of kernels increases, we begin to see asymmetry and direction sensitive features appear. In fig.~\ref{fig:multiE}, we show a comparison of the learned kernels when learning 1, 2, and 4 types of RGC events for interpolation with a bandwidth of roughly $150,000$ events per bin, along with the generated events for three different scenes. Learning one type of event, $RGC_1$, achieves an average PSNR of $36.52$dB. Learning just one more kernel, $RGC_2$, increases the performance to $36.92$dB. With four types of RGCs, $RGC_4$, we achieve $37.27$dB at the same bandwidth. Similarly, in our supplement, we show that at a lower total bandwidth of $10,000$ events per bin, $RGC_1$ achieves $33.76$dB, while $RGC_{16}$ achieves $35.16$dB. 

\section{Discussion and Conclusion}\label{sec:conclusion}
We present a biologically inspired way of sensing by creating a new event-generation framework that extends the traditional pixel-wise event paradigm by leveraging local spatial information. For this purpose, we craft a differentiable event simulator to enable learning of the RGC kernels. We demonstrate experiments on video interpolation and optical flow, showing that our learned events outperform both conventional DVS and center-surround DVS (CSDVS) methods under the same bandwidth constraints. By utilizing neighborhood context, our proposed approach delivers richer, more informative event streams—pointing to a promising new direction in event-based sensing for real-time and resource-constrained applications. 

\noindent\textbf{Hardware Feasibility.} 
There are several avenues for creating Neural Ganglion Sensors in hardware, especially with the rise of in-pixel compute. \cite{Delbruck2022-csdvs} explored the feasibility of a center-surround DVS and proposed using polysilicon lateral resistors to weight the surround according to their hand-crafted kernel. A similar resistor mesh could be applied with our learned kernel weights. For more adaptability where the weights can be dynamically updated on-the-fly, emerging in-pixel compute platforms such as~\cite{SCAMP5} can also be an attractive candidate. These emerging platforms integrate compute and memory into the sensor plane and have already shown promise in a number of computational imaging pipelines~\cite{neuralsensors, mantissacam, nguyen2022learning}. Implementing the ideal learned RGC kernels requires a multiplication in floating point precision. However, the operations available on in-pixel processors are currently limited. Some approximations or training with additional constraints would be required for implementation on the current generation of sensor--processors, such as operating in highly quantized regimes where kernels are binary or ternary~\cite{hmso2024pixelrnn, camthatcnns, bose_2020, Liu2020HighspeedLC}. As the capabilities of in-pixel compute continue to develop, our full floating point RGC kernels will quickly become feasible.

\noindent\textbf{Future Directions.} There are a number of potential extensions to this work. In particular, exploring non-binary nonlinearities could be beneficial to the vision tasks. Thresholding is currently used as it best aligns with what we see in existing event sensors, and the binary spikes are akin to the spiking potential in human retinas. \\

As the demand for efficient sensing grows across domains ranging from AR/VR headsets to drones and as emerging cameras place more compute in-pixel, we envision that these insights will help guide sensor designers in optimizing future event-sensing technologies. As compute capabilities and memory continue to improve in emerging sensor-–processors, Neural Ganglion Sensors will offer a promising balance— a small amount of compute on-sensor for massive bandwidth savings.

\ifpeerreview \else
\section*{Acknowledgments}
This project was in part supported by Samsung, SK Hynix, and the NSF Graduate Research Fellowship Program.
\fi

\bibliographystyle{IEEEtran}

\ifpeerreview \else

\vfill

\fi

\newpage

\setcounter{figure}{0}
\setcounter{section}{0}
\setcounter{equation}{0}
\clearpage
\twocolumn[%
\begin{center}
\section*{\Large  Neural Ganglion Sensors:\\Supplemental Material}
\end{center}
]
\renewcommand{\thefigure}{S\arabic{figure}}
\renewcommand{\thesection}{S\arabic{section}}
\renewcommand{\thetable}{S\arabic{table}}

\section{Additional Experimental Details }\label{sec:supp_baselines}

\subsection{Closed Form Binning, with refractory period}
In the main paper, we show the closed form equations to extract binned events from high speed video. Here, we extend them to account for the refractory period, $r$, the ``off'' period of a pixel after it has fired an event, and $p_r$, the time since the last event timestamp, $t^e_{mem}$, during event generation from the previous frame.
\begin{align}
p_r &= t_k - t^e_{mem}\\
    \alpha^r &= \left(r//\alpha + 1\right)\cdot \alpha \\
    \begin{split}
            \beta^r &= \beta+\text{max}(0, ((r-p_r)//\alpha) \cdot \alpha) \\
    &= \text{max}(0, ((r-p_r)//\alpha) \cdot \alpha) 
    \end{split} 
\end{align}
The weight and number of events become
\begin{align}
    w_i^r &= \frac{\beta^r + i*\alpha^r}{\text{bin}^+ - \text{bin}^-}\\
    \mathcal{N}^r &= (t_{k+1}-t_k -\beta^r)//\alpha_r
\end{align}
and the closed form $bin^-$ and $bin^+$ are
\begin{align}
\begin{split}
    \text{bin}^{-}_{frame} &= pol\cdot \left( 1 - \frac{\beta^r}{\text{bin}^+ - \text{bin}^-} \right) \cdot \mathcal{N}^r  \\
    &- \left(\frac{\alpha^r}{\text{bin}^+ - \text{bin}^-} \right) \cdot pol\cdot \frac{(\mathcal{N}^r+1)(\mathcal{N}^r)}{2}
\end{split}\\
\begin{split}
    \text{bin}^{+}_{frame} &= pol\cdot \left( \frac{\beta^r}{\text{bin}^+ - \text{bin}^-} \right) \cdot \mathcal{N}^r  \\
    &+ \left(\frac{\alpha^r}{\text{bin}^+ - \text{bin}^-} \right) \cdot pol\cdot \frac{(\mathcal{N}^r+1)(\mathcal{N}^r)}{2}
\end{split}
\end{align}

Now with these closed form equations and straight-through gradient estimation, we can model the gradients to enable learning even with non-zero refractory periods. In our experiments, the refractory period was set to 1 millisecond, following V2E2V~\cite{liu_sensing_2023}. We also model noise non-idealities with Gaussian noise with sigma $0.03$, following the V2E emulator~\cite{Hu2021v2e}. 
Ultimately, our closed form solution with gradients produces event bins that match closely with current state-of-the-art non-differentiable video to binned event pipelines.

\subsection{Dataset Considerations and Details}
For video interpolation, we utilize the GoPRO dataset~\cite{Nah_2017_CVPR}. High speed video frames are captured at 240 fps. There are 22 scenes in the train set and 11 scenes in the test set. To train, each sample is a 9-frame sequential subset of a scene. The two end-point frames are used as input to the base model, REFID, while the rest are used to simulate the events. The seven frames in-between the end-points are also used as the ground truth frames for comparison and metric calculation.  

For optical flow, we utilize the TartanAir Dataset: AirSim Simulation Dataset for Simultaneous Localization and Mapping ~\cite{tartanair2020iros}. In our experiments, we test our methods with the `Hard' subset of the TartanAir Dataset. There are 18 environments total. We randomly set `office2' and `westerndesert' scenes as validation, and `gascola' and `japanesealley' as the test scenes prior to any experiments. As mentioned in the main paper, in order to simulate events, we first interpolate 15 frames between each original neighboring frames. Similar to previous work~\cite{yang2024eventcameradatadense}, we generate the dataset of high speed video using EMA-VFI to interpolate for training. Our resulting training dataset consists of random crops from the original dataset along with the interpolated frames, creating a dataset of 44,776 training, 5,000 validation, and 5,000 test sequences. All experiments shown in this work are trained and evaluated on this created dataset. 
TartanAir has ground truth masks for optical flow which mask out pixels that have become occluded or disoccluded. Following the convention of previous work, we train and evaluate on the masked optical flow.

\subsection{Model Details}
\subsubsection{Event-based Video Interpolation}
We utilize REFID~\cite{sun2023event}, a state-of-the-art model for event-based video interpolation. It uses an event-guided adaptive channel attention and bidirectional event recurrent blocks. The model we use is the default from the repo with num-encoders=3, 32 base channels, num-block=1, and 2 residual blocks. With one type of RGC kernel, the default number of event channels is 2. As we experiment with different number of types of RGC kernels, we scale just the input layer accordingly. In all experiments, models are trained for 200k iterations using the PyTorch AdamW optimizer with a learning rate of $2e^{-4}$ for the interpolation model, $5e^{-5}$ for the RGC kernels, and weight decay $1e^{-4}$. For more details, see the code.

\subsubsection{Event-based Optical Flow}
We use IDNet~\cite{Wu_2024_ICRA}, the light-weight state-of-the-art model for event-based optical flow. The backbone is the recurrent neural network ConvGRU that processes the event bins sequentially. Specifically, we use IDNet's id-8x variation.  In all experiments, we train for 400k iterations with the PyTorch Adam optimizer with a learning rate of $1e^{-4}$.

\subsection{Additional Interpolation Results}
\begin{figure}
    \centering
    \includegraphics[trim={0 7cm, 12.5cm, 0}, clip, width=\linewidth]{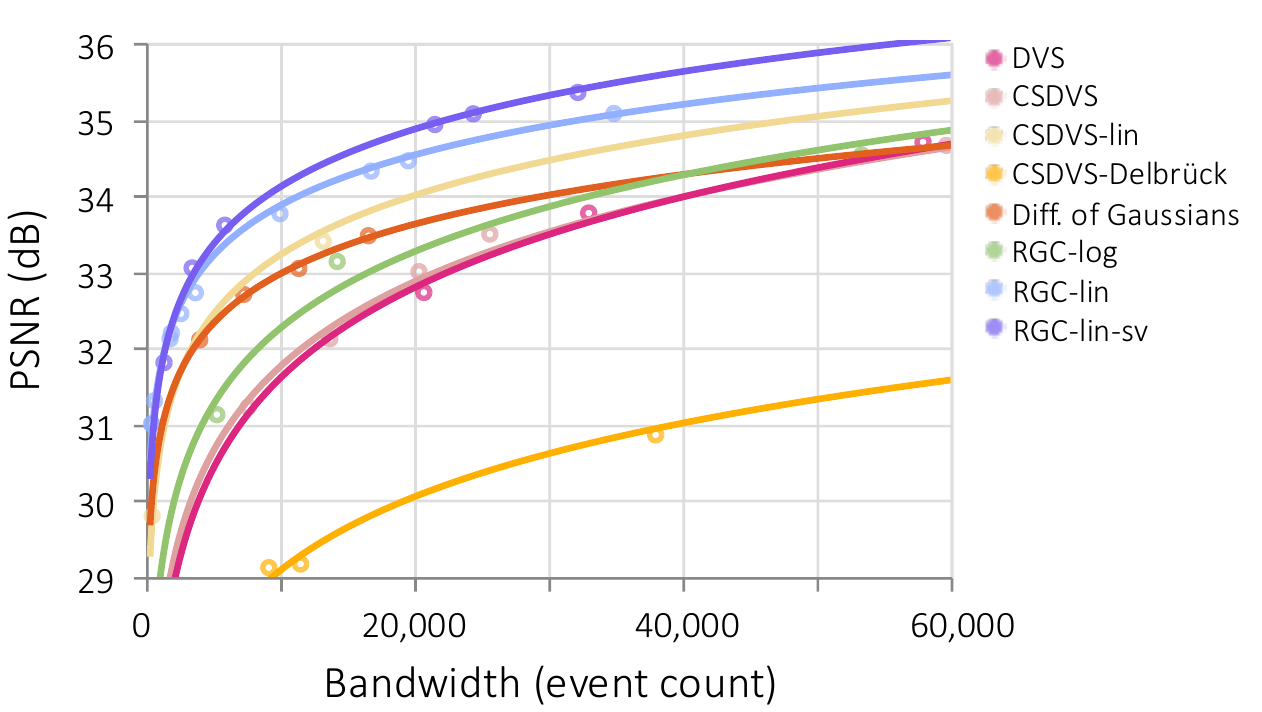}
    \caption{\textbf{Extended Comparisons for Video Interpolation: } Along with the comparisons in the main paper, we include the CSDVS-  in the linear intensity and the difference of Gaussians.}
    \label{fig:supp-interpplot}
\end{figure}
We include an extended comparison for bandwidth vs performance in fig. ~\ref{fig:supp-interpplot}. For a no-events baseline, we train the same model with zero events. The average PSNR over the test set is $28.91$dB and the SSIM is $0.912$. In addition to the CSDVS and CSDVS-Delbr\"{u}ck shown in the main paper, we explore CSDVS-lin, CSDVS in the linear intensity domain. In neuroscience, another common way to model the center-surround is with a Difference of two Gaussian kernels of different sizes. Here, we model the Difference of Gaussians, learning the standard deviation of the two Gaussians as we tune the bandwidth via the weighting on the sparsity loss. Fig.~\ref{fig:supp-interpplot} shows the full PSNR vs. Bandwidth trade-offs. We use a logarithmic fit for all event types. While CSDVS performs similarly to the DVS, CSDVS in the linear domain performs better. The Difference of Gaussians is also an improvement over the DVS kernel. While it may approximately model human RGCs well, for machine vision tasks, the best is still the learned RGC-lin and the spatially-varying RGC-lin-sv. 

\subsubsection{The effect of increasing the receptive field}
\begin{figure}
    \centering
    \includegraphics[trim={0 12cm 2cm 0},clip,width=\linewidth]{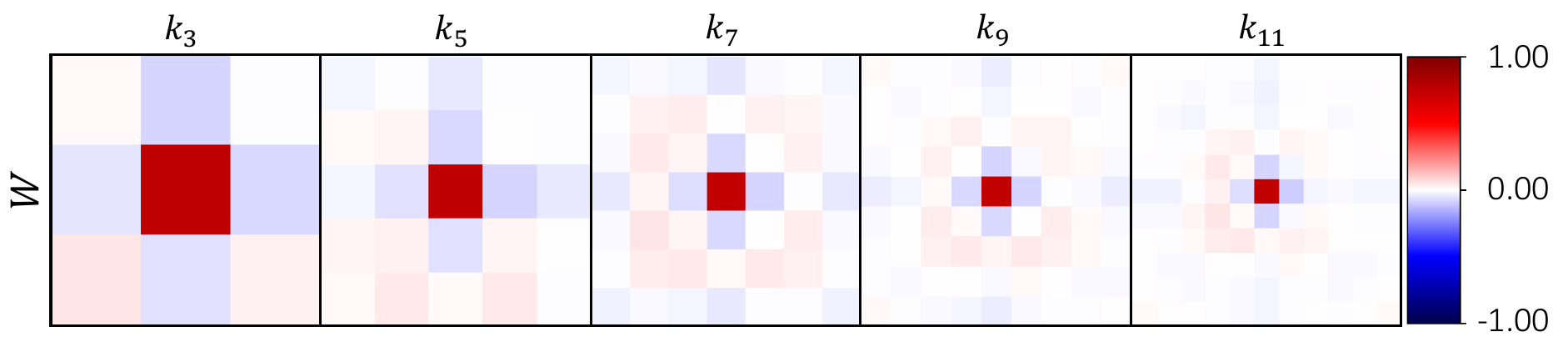}
    \vspace{-7mm}    \caption{\textbf{Learned kernels at increasing kernel sizes.} As our eye's RGCs have varying receptive field sizes, we test the effect of increasing the kernel for the task of interpolation, given a set bandwidth of $\approx20,000$ events per bin. As the kernel size increases, a subtle structure emerges where around the center pixel, it is negative, and then just further out is a ring of positive again before decreasing once more.  }
    \label{fig:kernel_sizes}
    \vspace{-3mm}
\end{figure}

We experiment with different kernel sizes $k=3,5,7,9,11$ for the video interpolation task. All of these models are trained with the same settings, except for kernel size. All models converged at roughly $20,000$ events per bin with a PSNR in the range $34.36$dB to $34.52$dB and SSIM of 0.97 across the board. Although the interpolation performance does not increase significantly as we increase the receptive field, interestingly a faint structure is revealed. Immediately outside of the center pixel, we have a negative weighting, and just further out, a positively weighted ring, before decreasing once more.

\subsubsection{The effect of increasing bandwidth}
\begin{figure}
    \centering
    \includegraphics[trim={0 12cm 2cm 0},clip,width=\linewidth]{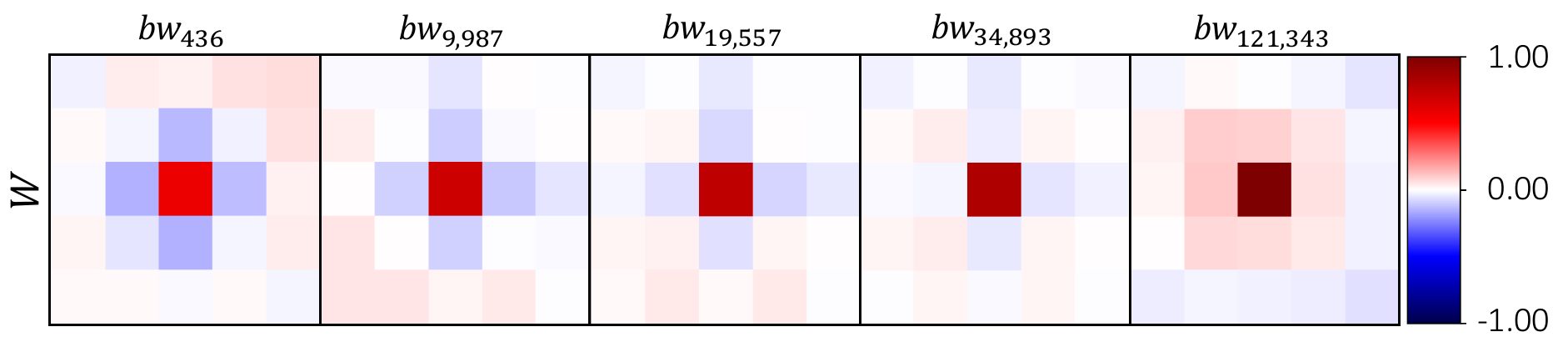}
    \vspace{-7mm}    \caption{\textbf{Learned kernels at increasing bandwidth.}  We optimize performance while tuning the bandwidth jointly by varying the weighting of the sparsity loss during training. Here we show the learned $5\times5$ kernels for different bandwidths (bw). }
    \label{fig:kernel_bw}
    \vspace{-3.5mm}
\end{figure}
As we tune the bandwidth by weighting the sparsity loss more or less, the learned kernel also changes. In fig.~\ref{fig:kernel_bw}, we show the learned RGC kernel for models trained at different bandwidths. The performance increases from $31.01$dB with $436$ events per bin to $34.45$dB with $19,557$ events, to $36.52$dB with $121,343$ events.

\subsection{Multi-channel RGCs for Video Interpolation}

\begin{figure*}
    \centering
    \includegraphics[trim={0 3.5cm 2cm 0cm}, clip, width=\linewidth]{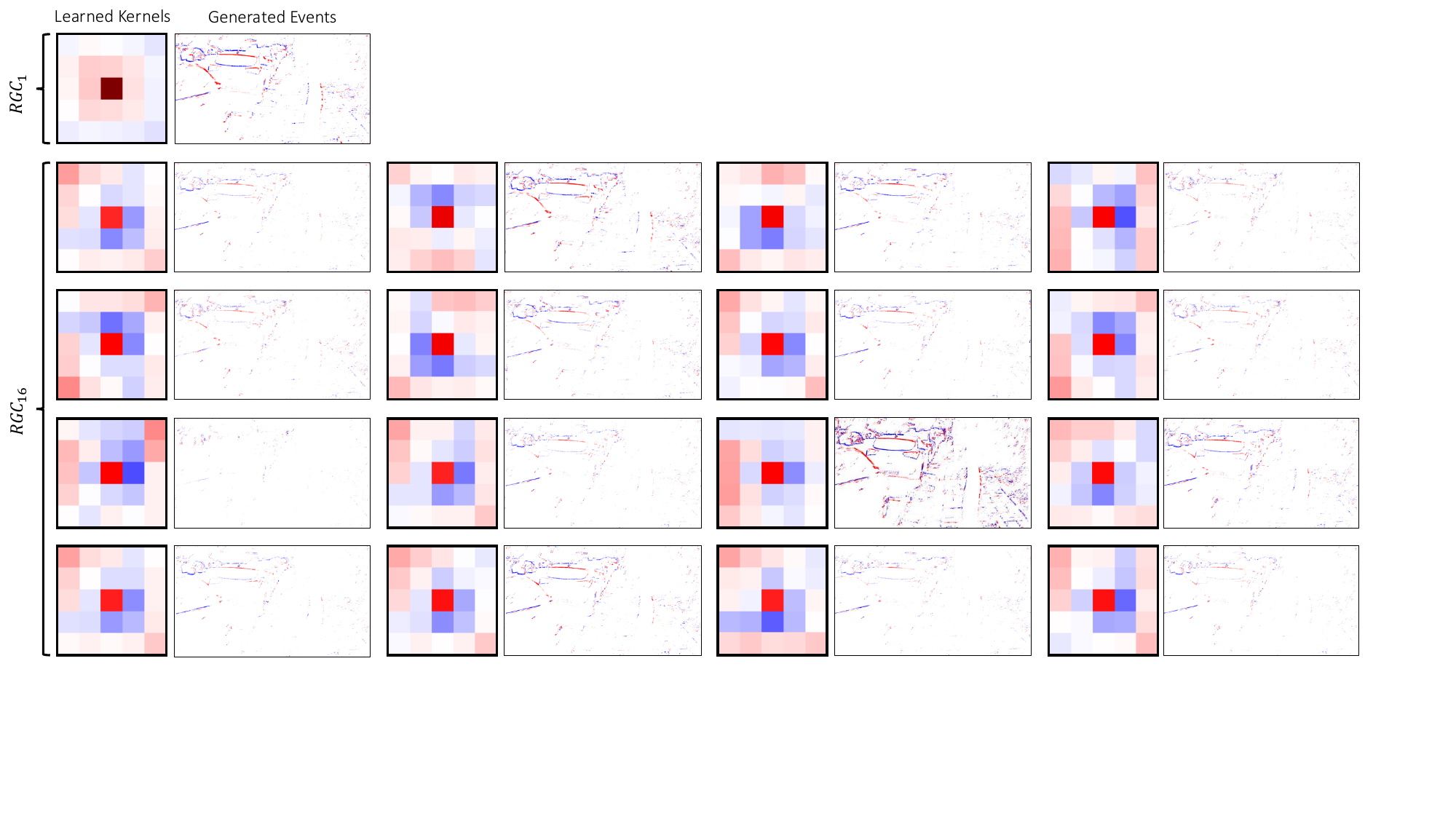}
    \\
    \includegraphics[trim={0 3.5cm 2cm 0cm}, clip, width=\linewidth]{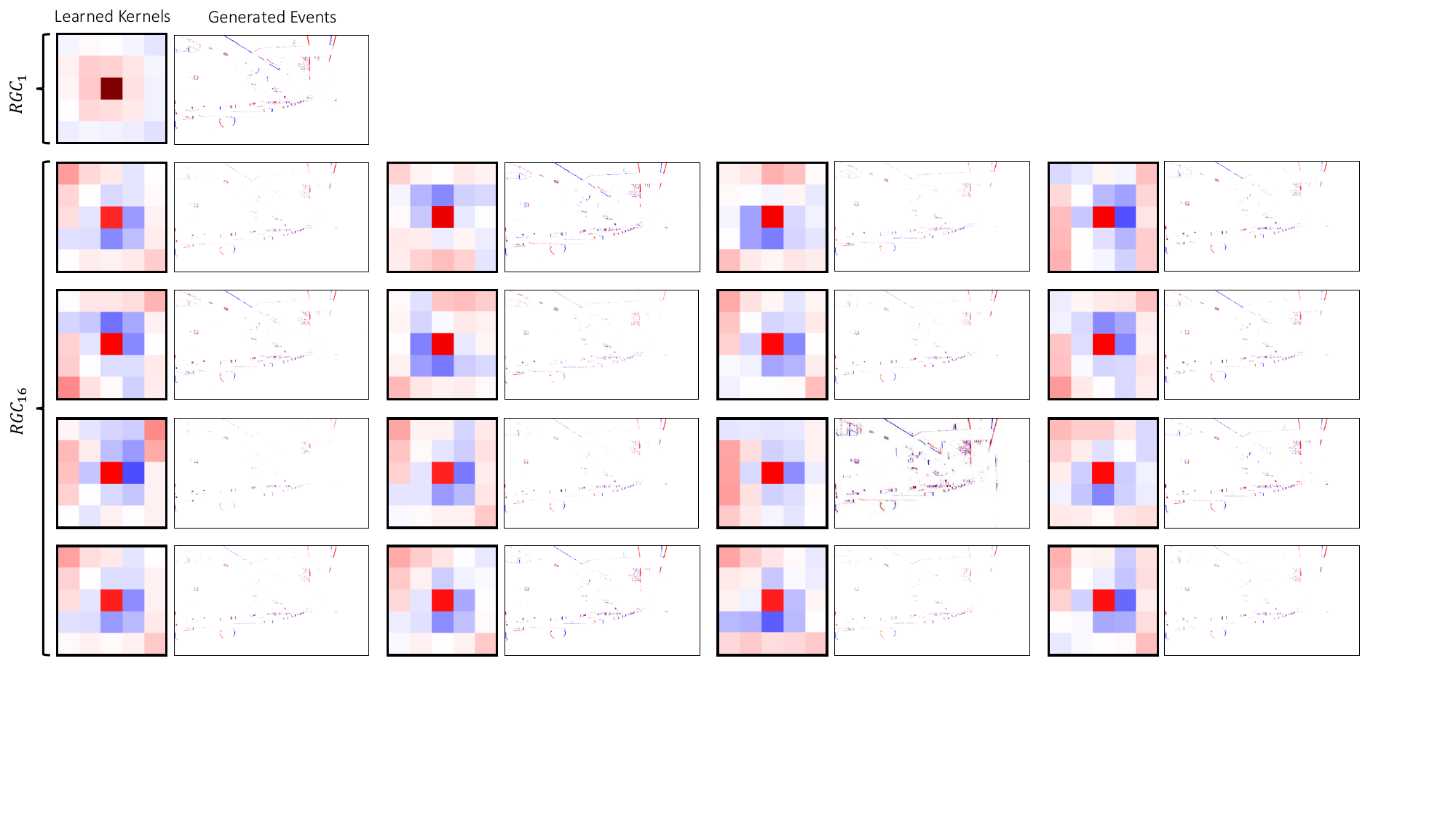}
    \caption{\textbf{Multi-channel Comparison:} Above are the learned kernels for $K_1$ and $K_{16}$ along with the events generated by the kernels for two scenes.}
    \label{fig:multi-16}
    \vspace{-3mm}
\end{figure*}

\begin{figure*}
    \includegraphics[trim={0 1.5cm 0 0cm}, clip, width=\linewidth]{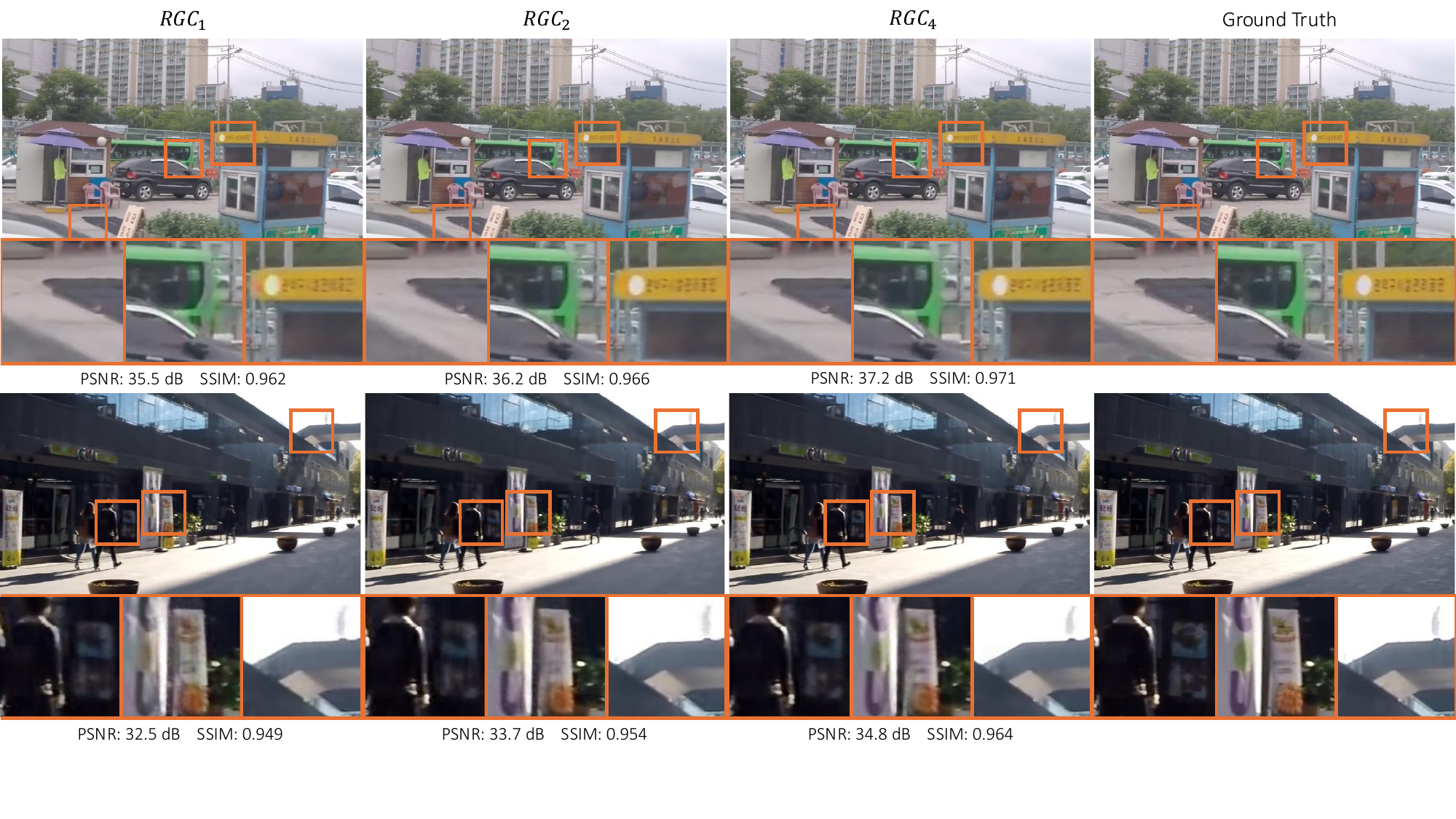}
    \\
    \includegraphics[trim={0 2cm 0 0.8cm}, clip, width=\linewidth]{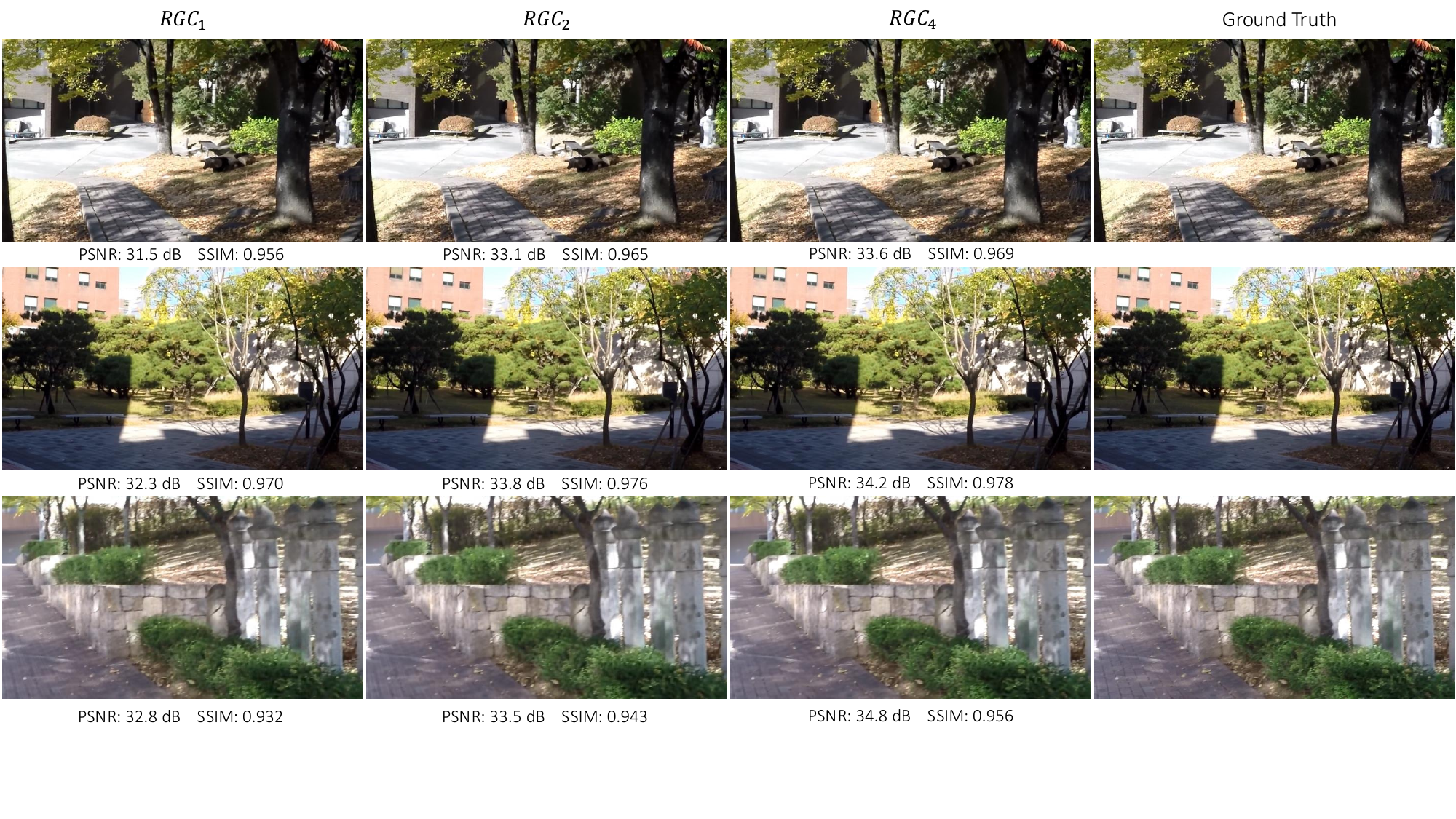}
    \caption{\textbf{Interpolation reconstructions from $RGC_1$, $RGC_{2}$, and $RGC_{4}$.} We show the middle reconstructed frame with PSNR and SSIM metrics.}
    \label{fig:k1k2k4_supp}
    \vspace{-5mm}
\end{figure*}

\begin{figure*}
\includegraphics[trim={0 7cm 0 0}, clip, width=\linewidth]{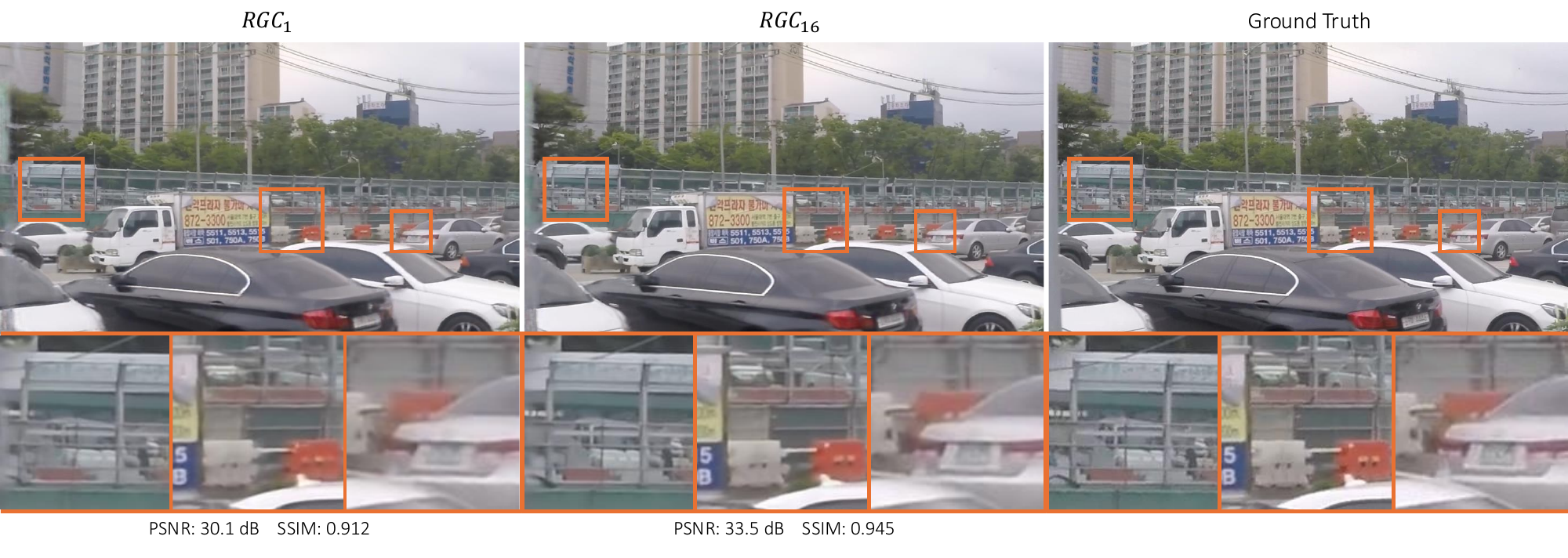}
\\
\includegraphics[trim={0 0cm 0 0.8cm}, clip,width=\linewidth]{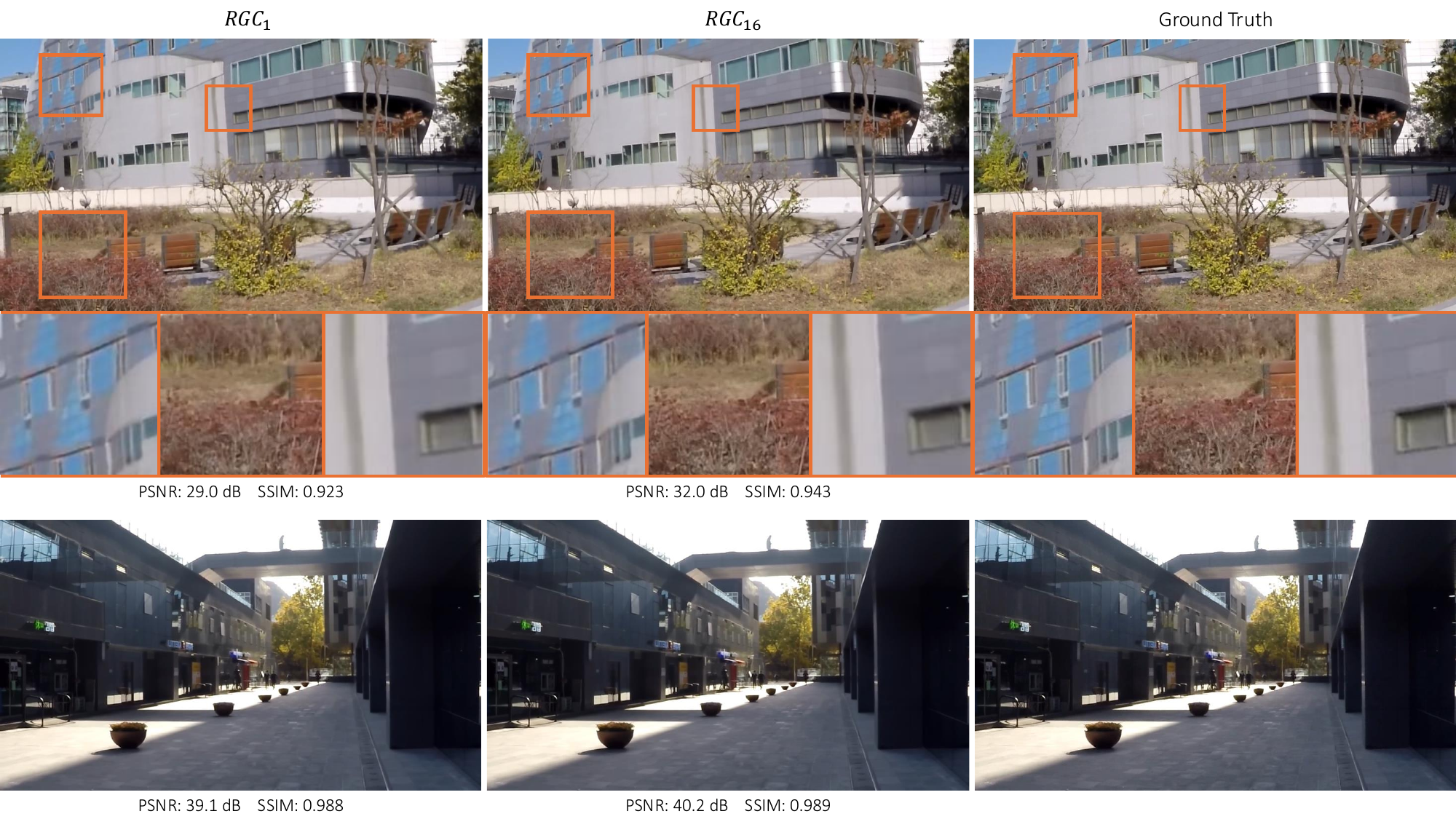}
\caption{\textbf{Interpolation reconstructions from $RGC_1$ and $RGC_{16}$:} We show the middle reconstructed frame for the one learned kernel and 16 learned kernel models trained at the ultra-low bandwidth of roughly $10,000$ events. }
\label{supp-fig:K1_K16_interp}
\end{figure*}
\subsubsection{Additional Multi-channel Results}
In the main paper, we show the comparison of learning 1, 2, and 4 different, yet complementary, kinds of RGCs. Here, we show qualitative results for learning 16 kernels. Fig.~\ref{fig:multi-16} provides a view of the 16 learned kernels and the single learned kernel comparison at the ultra-low-bandwidth regime of roughly 10,000 events. We show two scenes and the events generated by each kernel. While the kernel learned for single RGC is quite symmetric, we can see the 16 kernels model can learn more specific features. We show samples of the reconstruction by the models with different number of channels in fig. ~\ref{fig:k1k2k4_supp} and fig. ~\ref{supp-fig:K1_K16_interp} with PSNR and SSIM metrics and some with zoom-in details. With more learned kernels, there is less warping and color discrepancies.

\subsubsection{Trade-off of Performance and Bandwidth}
\begin{figure}[h]
    \centering
    \includegraphics[trim={0 7cm, 13.5cm, 0}, clip, width=\linewidth]{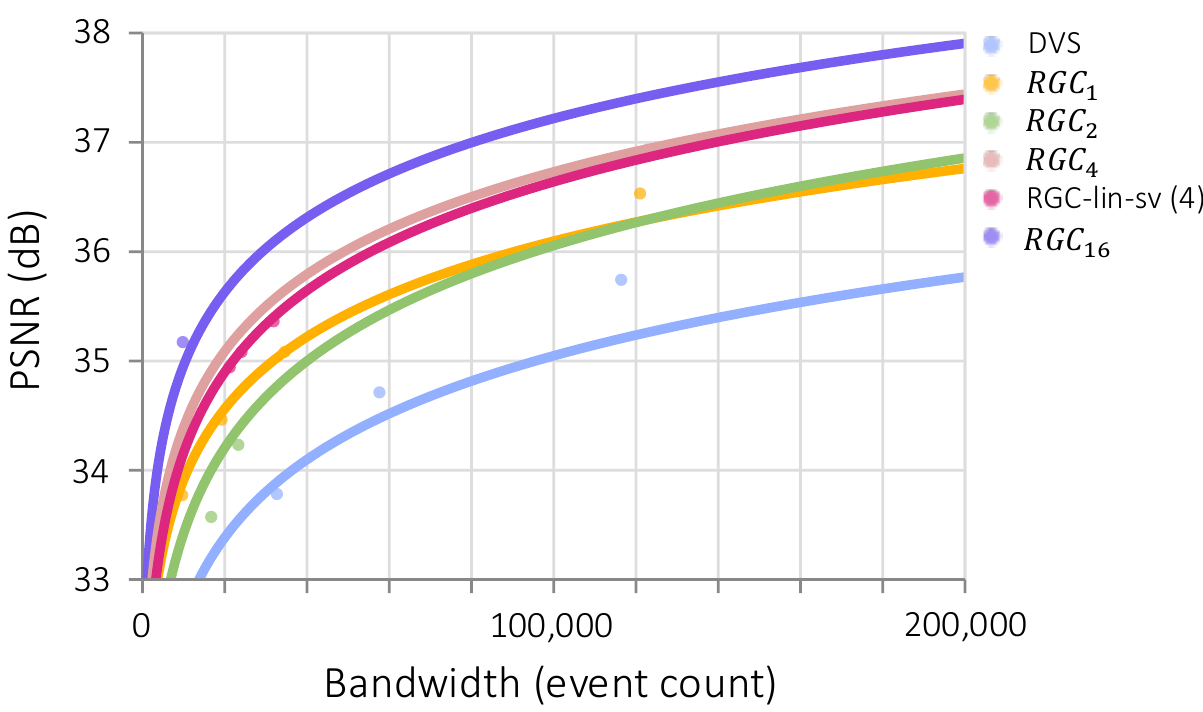}
    \caption{\textbf{Multi-channel RGCs for Video Interpolation: } We explore the idea of having mulitple learned retinal circuits per pixel, similar to the the human retina. $K1$ is a single kind of event, $K2$ is two kinds of events, and so on. In the main paper, we also had the option to learn spatially-varying kernels. This means there are 4 unique kernels, but each pixel only gets one circuit. We show this setting, RGC-lin-sv (4) as a comparison. K4 and RGC-lin-sv perform similarly. We also show DVS for reference.}
    \label{fig:supp-multievent_plot}
\end{figure}

Fig. ~\ref{fig:supp-multievent_plot} 
shows how using multiple kinds of events can further improve the performance at any given bandwidth. The $i$ in $RGC_{i}$ refers to how many parallel circuits are learned per pixel. This is similar to how our eyes operate. For a given receptive field, there can be multiple kinds of RGCs looking at the signals. In fig. ~\ref{fig:supp-multievent_plot}, we also show DVS for reference as well as the RGC-lin-sv model, which is spatially varying kernels. $RGC_4$ and RGC-lin-sv are similar in that they both have 4 learned kernels. However, the spatially varying model is 1 kernel per pixel, as opposed to 4 per pixel. We can see that the performance for video interpolation is similar.

\subsubsection{Additional considerations for Multi-event Bandwidth}
Typically, the event packet that the traditional DVS outputs includes the pixel location and a bit denoting the polarity. In the multi-event case, we can send the location and a few bits. As we see in the generated events by each kernel in the main paper, one or two types of RGC seem to dominate or produce the most events, while the others complement with additional information. There can be an improvement in bandwidth if the system can support variable length packets. We can take advantage of Huffman encoding where fixed binary codes can be assigned to different RGC types based on their expected frequency. In this way, it can send the minimum number of bits and only send additional bits for rarer events when needed. Alternatively, when learning only a few RGC kernels, using spatially-varying kernels as demonstrated with RGC-lin-sv could potentially enable most of the performance benefits without requiring additional bits to be read out.

\subsection{Additional Optical Flow Samples}
\begin{figure}[h]
    \centering
    \includegraphics[width=\linewidth]{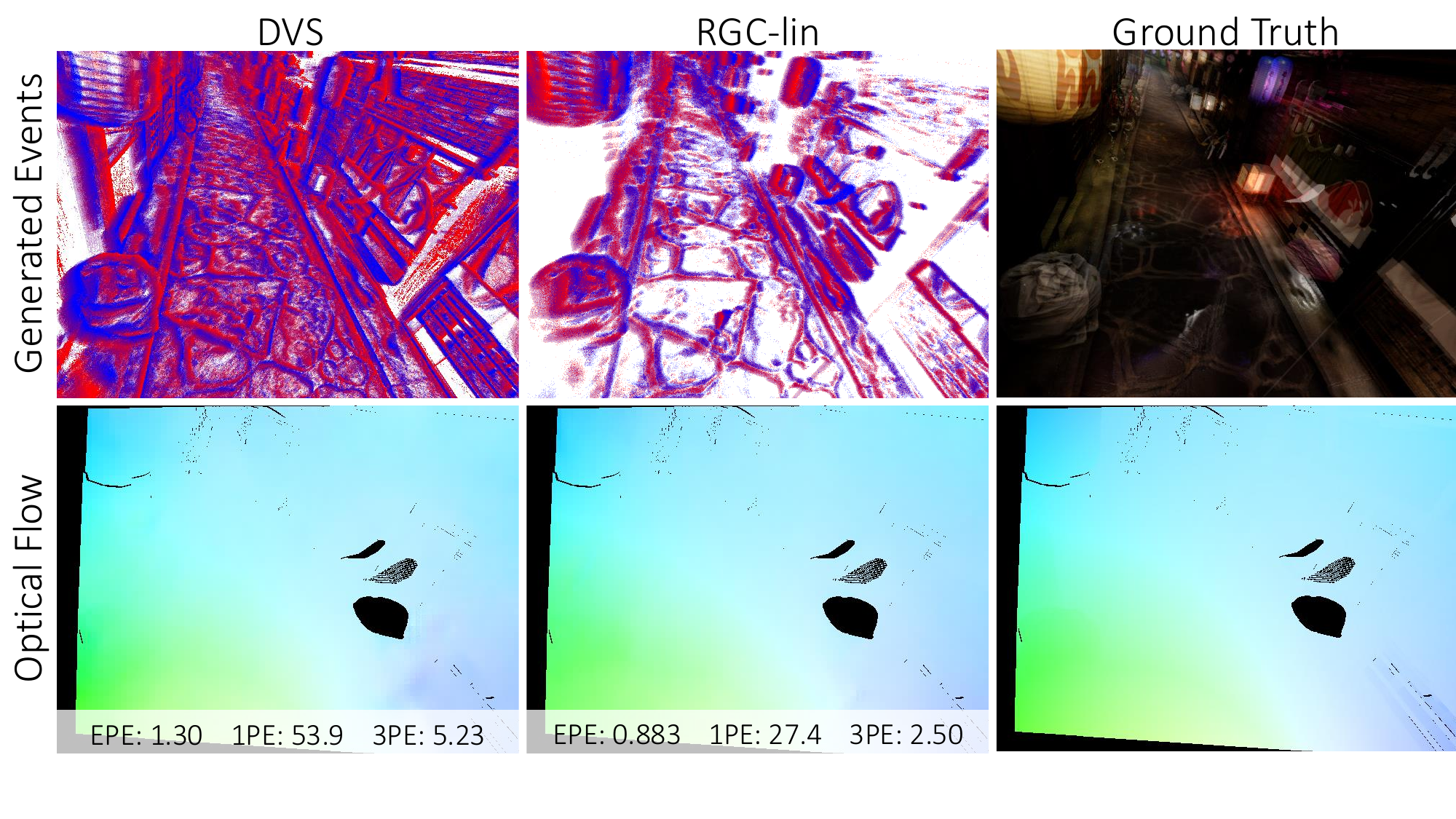}
    \\
    \vspace{-2mm}
    \includegraphics[width=\linewidth]{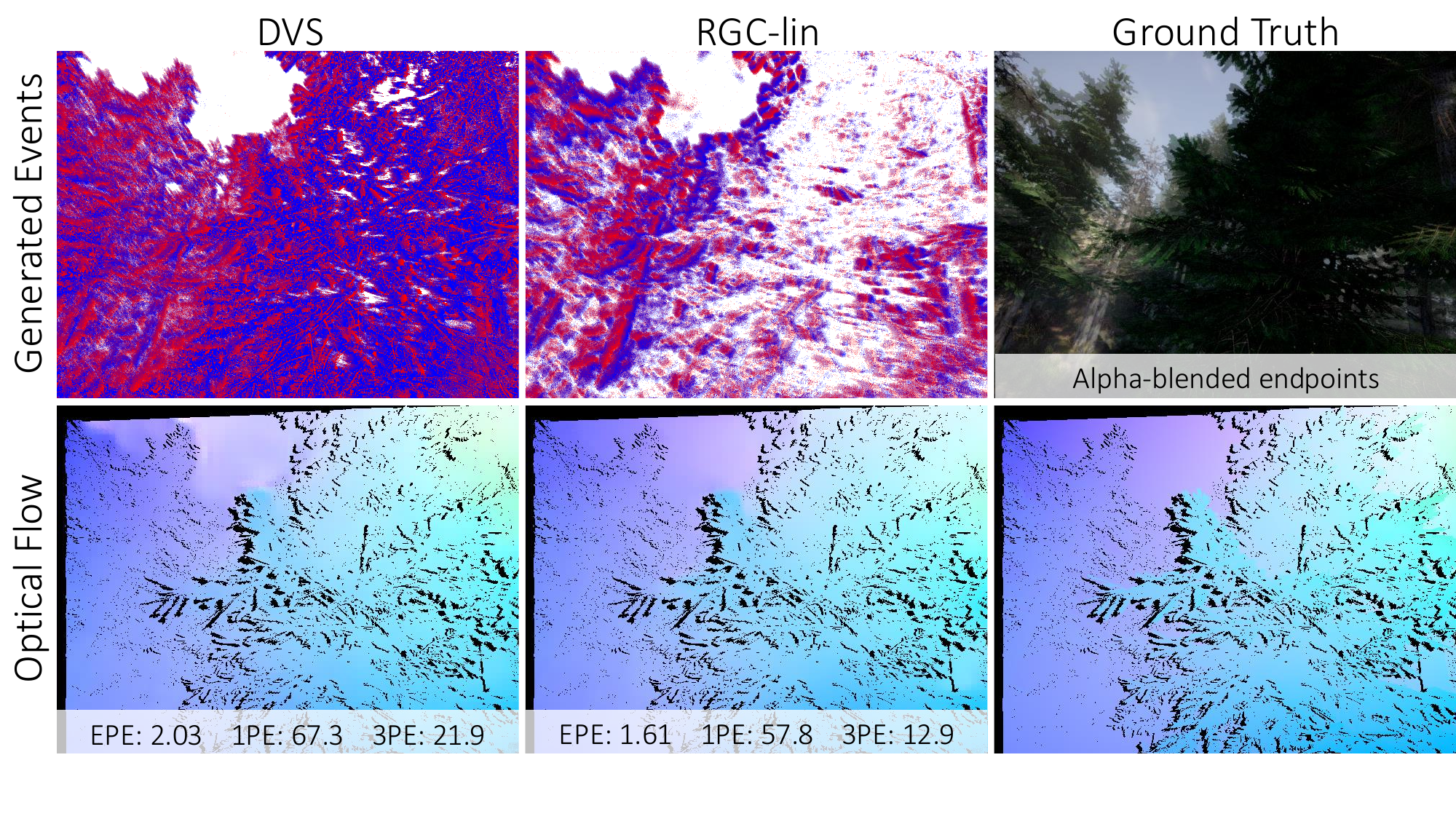}
    \\
    \vspace{-2mm}
    \includegraphics[width=\linewidth]{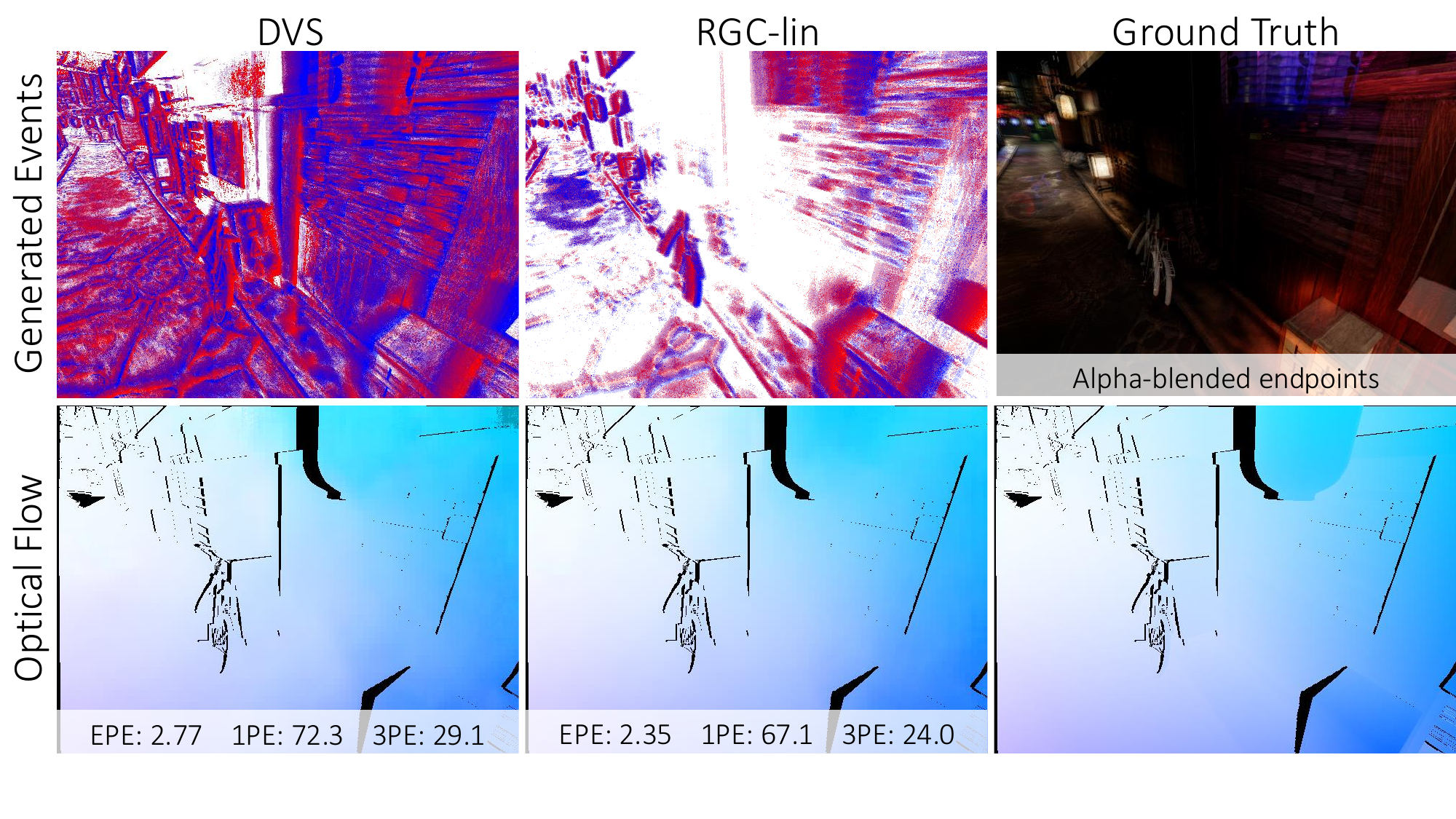}
    \\
    \vspace{-2mm}
    \includegraphics[width=\linewidth]{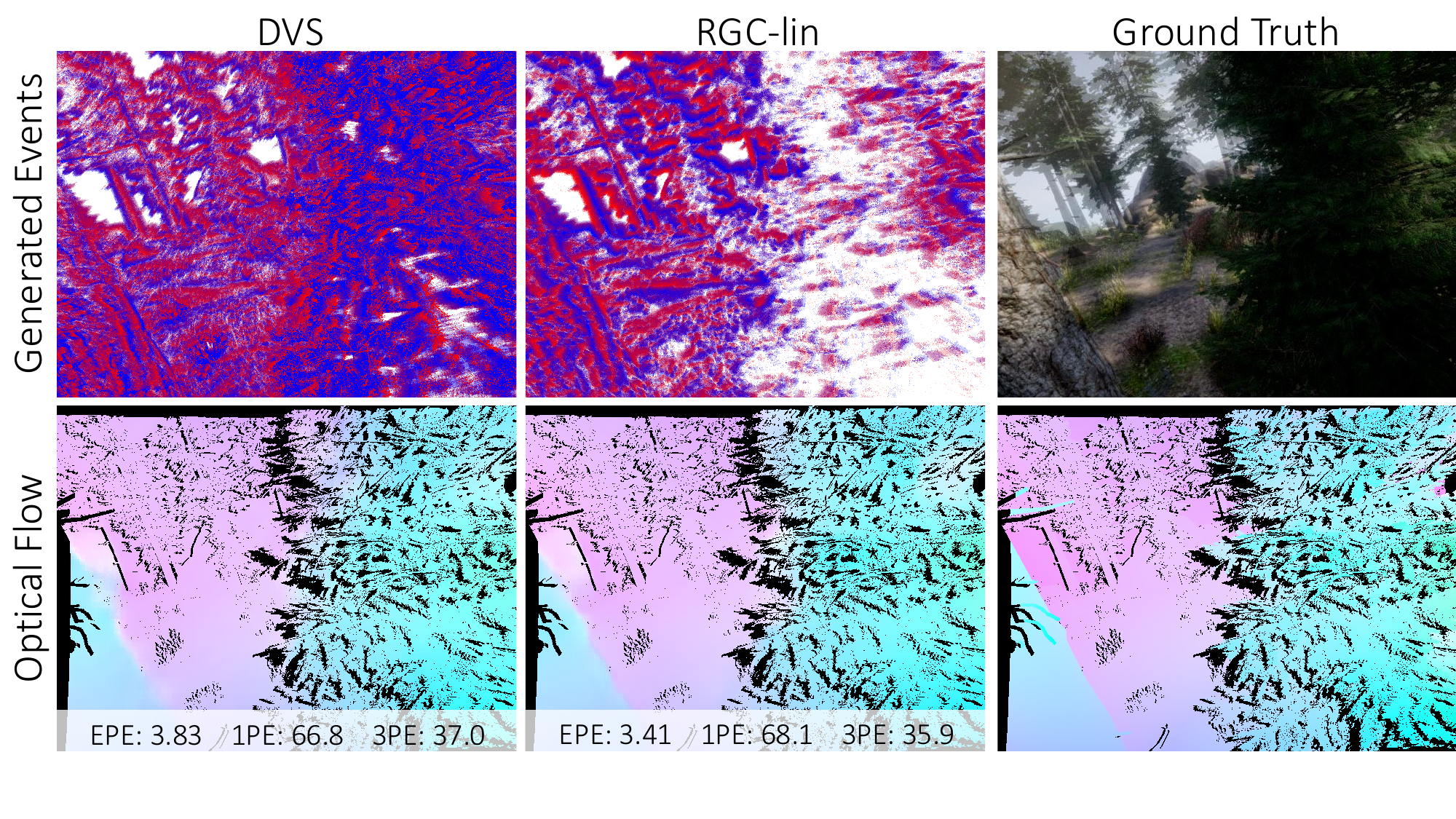}
    \vspace{-9mm}
    \caption{\textbf{Additional Optical Flow Qualitative Results.} We show EPE$\downarrow$, 1PE$\downarrow$, and 3PE$\downarrow$ metrics for four additional scenes. With our learned RGC kernel, we achieve both sparser readout and better optical flow performance.}
    \label{fig:supp_flow_results}
    \vspace{-5mm}
\end{figure}

We provide more qualitative results for optical flow in fig. ~\ref{fig:supp_flow_results}. Again, we compared the DVS model with the best performance, DVS 0.1T, against our learned, single kernel model. Even with half the average number of events, we achieve better optical flow.

\end{document}